\documentclass[final]{cvpr}

\usepackage{times}
\usepackage{epsfig}
\usepackage{graphicx}
\usepackage{amsmath}
\usepackage{amssymb}
\usepackage{enumitem}

\usepackage{subcaption}
\usepackage{booktabs}
\usepackage{gensymb}
\usepackage[ruled,vlined]{algorithm2e}
\usepackage{multirow}
\usepackage{makecell}
\usepackage[table, svgnames, dvipsnames]{xcolor}
\usepackage[pagebackref=true,breaklinks=true,colorlinks,bookmarks=false]{hyperref}

\def\cvprPaperID{3645} 
\def\confYear{CVPR 2021}

\begin{document}

\title{Topological Planning with Transformers for Vision-and-Language Navigation}

\author{Kevin Chen\\
Stanford University\\
{\tt\small kevin.chen@cs.stanford.edu}
\and
Junshen K. Chen\\
Stanford University\\
{\tt\small jkc1@stanford.edu}
\and
Jo Chuang\\
Stanford University\\
{\tt\small jochuang@stanford.edu}
\and
Marynel Vázquez\\
Yale University\\
{\tt\small marynel.vazquez@yale.edu}
\and
Silvio Savarese\\
Stanford University\\
{\tt\small ssilvio@stanford.edu}
}


\maketitle

\begin{abstract}
   Conventional approaches to vision-and-language navigation (VLN) are trained end-to-end but struggle to perform well in freely traversable environments. Inspired by the robotics community, we propose a modular approach to VLN using topological maps. Given a natural language instruction and topological map, our approach leverages attention mechanisms to predict a navigation plan in the map. The plan is then executed with low-level actions (e.g. {\sc forward}, {\sc rotate}) using a robust controller. Experiments show that our method outperforms previous end-to-end approaches, generates interpretable navigation plans, and exhibits intelligent behaviors such as backtracking.
\end{abstract}

\begin{figure*}[t!p]
\centering
\includegraphics[width=\linewidth]{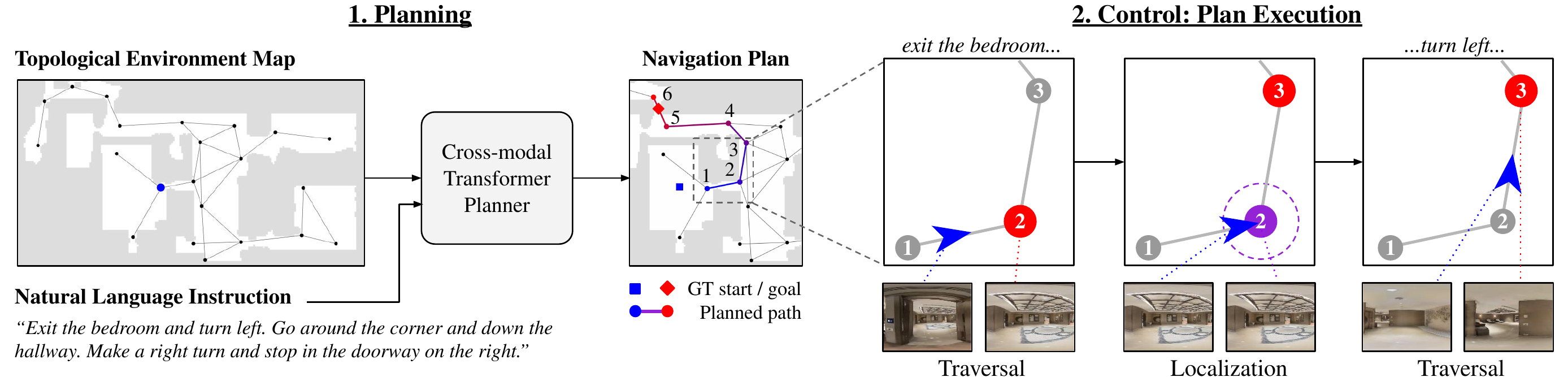}
\caption{The agent uses the natural language instruction to generate a navigation plan in the topological map (left). A controller then executes the predicted plan by sequentially traversing to each subgoal node in the plan (right). As the agent approaches the subgoal node, it consumes that active node, and the subsequent node in the plan becomes the new active node.}
\label{fig:pull_figure}
\end{figure*}

\section{Introduction}

Enabling robots to understand natural language and carry out communicated tasks has long been desired. A critical step towards this goal in the context of mobile robotics is vision-and-language navigation (VLN) \cite{anderson2018vision}. In VLN, the agent is provided with a navigation instruction such as: ``\textit{Exit the bedroom, walk to the end of the hall, and enter the kitchen on your right}.'' The agent is then expected to follow the instruction and move to the specified destination.

The majority of VLN systems \cite{fried_nips2018,hao2020towards,ke2019tactical,ma2019selfmonitoring,ma2019regretful,tan2019learning,wang2019reinforced,zhu2020vision} are end-to-end deep learning models and utilize unstructured memory such as LSTM~\cite{hochreiter1997long}. These methods work well when the movement is constrained to pre-defined locations, but performance drops significantly when the agent is allowed to move freely \cite{krantz2020navgraph}. Moreover, learning to perform navigation, including mapping, planning, and control, in a fully end-to-end manner can be difficult and expensive. Such approaches often require millions of frames of experience \cite{chen2019learning,wahid2020learning,wijmans2019dd}, and yet performance  substantially degrades without ground truth odometry \cite{chen2019learning,wijmans2019dd}. 

To address the aforementioned issues, recent visual robot navigation literature has explored using structured memory (e.g. metric maps, topological memory) and using a modular approach, where the algorithm is explicitly divided into relevant subcomponents such as mapping, planning, and control  \cite{chaplot2019learning,chaplot2020neural,Savarese-RSS-19,fang2019scene,gupta2017cognitive,karkus2019differentiable,meng2019scaling,SavinovDosovitskiyKoltun2018_SPTM}. These approaches have been demonstrated to work well for tasks such as target image navigation and environment exploration. However, they have not been well studied in the context of VLN.

In this work, we employ a modular approach and leverage topological maps for VLN. Topological maps, inspired in part by cognitive science, typically represent environments as graphs where nodes correspond to places and edges denote environment connectivity or reachability. Compared with metric maps, topological maps eliminate the need for meticulous map construction. They promote efficient planning, interpretable navigation plans, and navigation robustness using cheaper sensors \cite{meng2019scaling,thrun1998learning}. In particular, the symbolic nature of topological maps lends them suitable for navigation with language \cite{matuszek2010following,thrun1998learning}, as the space discretization provided by the maps can facilitate  learning a relationship between instructions and spatial locations.

Importantly, using topological maps synergizes well with sequence prediction models. Predicting a navigation plan in a topological map bears many similarities with predicting sequences for language tasks such as language modeling and neural machine translation (NMT). By drawing the parallel between navigation planning and language sequence prediction, we can leverage powerful attention mechanisms \cite{vaswani2017attention} that have enabled significant breakthroughs in language tasks for the navigation problem. Recently, these attention-based models have even been demonstrated to achieve comparable performance to convolutional neural networks on image recognition tasks \cite{dosovitskiy2016learning}.

We propose using a cross-modal attention-based transformer to compute navigation plans in topological maps based on language instructions. Whereas a language model predicts a word at a time, our transformer predicts one topological map node in the navigation plan at a time. Structuring our model in this manner allows the agent to attend to relevant portions of the navigation instruction and relevant spatial regions of the navigation environment during the navigation planning process. For example, this enables our model to relate the word ``bedroom'' in a navigation instruction to the physical room in the environment.


Altogether, we propose a full navigation system for VLN. Unlike much of the prior work in VLN, we use a more challenging setup, allowing the agent to freely traverse the environment using low-level discrete actions. To this end, we define a topological map representation that can be constructed by the agent after it has freely explored the environment. The maps are used as part of a modular navigation framework which decomposes the problem into planning and control (Fig. \ref{fig:pull_figure}). For each navigation episode, our agent first uses the cross-modal transformer to compute a global navigation plan from the navigation instruction and topological map. This navigation plan is executed by a robust local controller that outputs low-level discrete actions.



We evaluate our approach using the VLN-CE dataset \cite{krantz2020navgraph}. Our experiments show that cross-modal attention-based planning is effective, and that our 
modular approach enables learning a robust controller 
capable of 
correcting for navigation mistakes like 
moving in the wrong direction.
\section{Related Work}


\paragraph{Language modeling and pretraining.} Recent language model work has benefited from attention-based transformer architectures \cite{vaswani2017attention} and generalized pre-training of language models for fine-tuning on specific tasks \cite{devlin2018bert}. With enough scale, these pretrained models are adaptive to new tasks, in some cases not even requiring any fine-tuning or gradient updates \cite{brown2020language}. These advances tie into the increasing interest in more effective methods for image recognition \cite{dosovitskiy2020image} and language grounding with visual cues \cite{vilbert, Su2020VL-BERT:,tan-bansal-2019-lxmert}. In addition to capturing semantics that can only be visually demonstrated, these models allow for a more diverse range of applications than language-only approaches including VLN.

\paragraph{Vision-and-language navigation (VLN).} Much of the progress in VLN has been achieved using end-to-end models \cite{deng2020evolving,fried_nips2018,hao2020towards,ke2019tactical,ma2019selfmonitoring,ma2019regretful,tan2019learning,wang2019reinforced,zhu2020vision} trained using the Matterport3D Simulator~\cite{anderson2018vision} in which the agent assumes a panoramic action space \cite{fried_nips2018} that allows it teleport to and from a fixed set of pre-defined locations in the environment. This setup promotes fast iteration and evaluation of different VLN approaches. However, it ignores the problem of associating action directions in the panorama with the pre-defined positions in the environment as well as the motion feasibility problem of moving from one location to another. As pointed out with the end-to-end approach by Krantz \etal~\cite{krantz2020navgraph}, the VLN problem becomes significantly harder when the agent is allowed to freely traverse the environment. In contrast, in our work the agent builds a topological map of the environment from ground up using exploration trajectories. The agent predicts a navigation path in the map and uses the local controller to execute the predicted path with low-level discrete actions, bringing the task of VLN one step closer to reality.

Although they operate in the simplified action space, Hao \etal~\cite{hao2020towards} and Majumdar \etal~\cite{majumdar2020vlnbert} propose attention-based transformer approaches that perform self-supervised pre-training on a cross-modal network using visual language pairings, either from the web or from demonstrations of a typical VLN trajectory. By fine-tuning these pre-trained models on specific VLN tasks, they effectively achieve better generalized performance across different variants of navigation tasks. Their approaches operate on top of an existing navigation graph and perform cross-modal attention between image regions and language instead of spatial regions and language. Additionally, Majumdar \etal propose a trajectory scoring mechanism and relies on other methods to generate candidate routes. On the other hand, our approach predicts routes and executes them with low-level actions.

\paragraph{Memory for visual navigation.} Early learning-based approaches to visual navigation were reactive \cite{dosovitskiy2016learning,zhu2017target} or based on unstructured memory \cite{mirowski2016learning,mnih2016asynchronous}. Later works investigate explicit representations of environments, such as metric map-based representations \cite{chaplot2019learning,gupta2017cognitive,gupta2017unifying,gupta2019cognitive,parisotto2018neural,zhang2017neural}. For example, \cite{anderson2019chasing} use metric maps for VLN, and \cite{fang2019scene} use transformers for explorative tasks by storing observations and poses as memory. Topological maps have also been demonstrated across different navigation problems \cite{chaplot2020neural,Savarese-RSS-19,meng2019scaling,SavinovDosovitskiyKoltun2018_SPTM}. For language navigation, \cite{matuszek2010following} and \cite{zang2018translating} use topological maps but do not use RGB cameras as in our work.

\section{Method}

\begin{figure}
     \centering
     \includegraphics[width=0.99\linewidth]{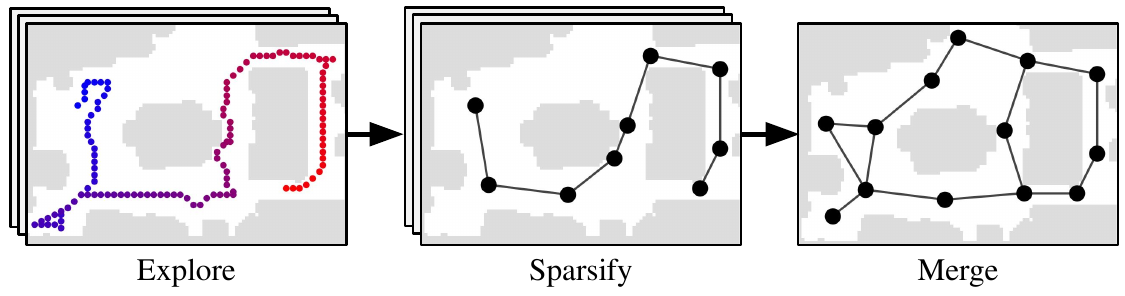}
     \caption{\textbf{Topological map construction.} The agent explores the environment multiple times, sparsifies the generated graph from each trajectory, and merges them together.}
    \label{fig:topological-map}
\end{figure}

\paragraph{Problem setup.} Our setup allows the agent to explore the environments using \textit{exploration trajectories} (Fig. \ref{fig:topological-map}) prior to executing the VLN tasks. This mimics settings closer to the real world in which the agent has to build an understanding of the indoor environment on its own rather than being handed a pre-defined map. After the exploration phase, the agent is expected to perform VLN and is provided with (1) the instruction text and (2) the current RGBD panorama observation. Our agent has orientation information (or heading) but not positional information during navigation time.


\paragraph{Overview of our approach.} From the exploration phase, the agent builds a topological map of the environment (Sec. \ref{sec:topomaps}). The agent then uses this topological understanding of the environment along with the instruction text and current observation to execute the VLN task.

For executing the VLN task, we take inspiration from the robotics community and use a modular approach. Specifically, we separate our approach into planning and control. Using a modular approach has been proven to work well \cite{nilsson1984shakey,thrun2006stanley} and has numerous advantages including interpretability, robustness, and flexibility \cite{karkus2019differentiable}.

In the planning stage, prior to taking any action steps in the environment, the agent first uses the navigation instruction to generate an \textit{interpretable} global navigation plan through the environment (Sec. \ref{sec:planner}). The robot then tries to follow the navigation plan using a repeating cycle of localization and control until it reaches the destination (Sec. \ref{sec:controller}). We use a hierarchical controller that first predicts waypoint subgoals. Then the waypoints are translated into low-level actions like {\sc forward} ($0.25m$) or {\sc rotate} (15\degree).

\subsection{Topological Map Representation}
\label{sec:topomaps}

Before planning can take place, the agent must first construct a topological representation of the environment. For VLN, it would be desirable that the map 1) covers a large proportion of traversable area; 2) contains rich information in nodes and edges to enable localization as well as planning from language; and 3) consists of connected nodes that are reachable with high probability when executing the plan.

Similar to prior work \cite{chaplot2020neural,meng2019scaling,SavinovDosovitskiyKoltun2018_SPTM}, we represent the environment as a graph in which each node is an observation and each edge represents the connectivity (or reachability) between two nodes. As the agent explores the environment, it places a node over its current location and connects this node to the previous one.

In prior work~\cite{meng2019scaling,SavinovDosovitskiyKoltun2018_SPTM} the agent may end up placing multiple nodes in the same location. For example, two observations from the same position can look very different if they are oriented $180\degree$ apart. This may result in dense topological maps and exacerbate difficulties in localization, planning, and control as each node only contains partial information about each location. To reduce redundancy and mitigate these issues, we represent the nodes as $360\degree$ panoramas oriented in a consistent fixed global orientation.

An overview of the topological map construction is presented in Fig.~\ref{fig:topological-map}. We run multiple pre-defined exploration trajectories per environment. The graph from each exploration trajectory is sparsified by a reachability estimator as proposed by Meng \etal~\cite{meng2019scaling} and then merged into a single graph that can be used for localization and planning. Further details on the exploration trajectories and map construction can be found in the appendix.

We append coarse geometric information to each directed edge in the map. To this end, we introduce a quantized polar coordinate system with 8 directions and 3 distances ($0$-$2m$, $2$-$5m$, $>5m$). Such information is useful for resolving ambiguities (e.g. symmetric environments) and may facilitate language/spatial reasoning. Since mapping is not our main focus, we use ground truth odometry to compute the edge categories during map construction. However, this assumption is relaxed during navigation test time and we instead use a neural network for localization (Sec. \ref{sec:controller}). 

\subsection{Cross-Modal Planning}
\label{sec:planner}


\begin{figure}
\begin{center}
   \includegraphics[width=0.75\linewidth]{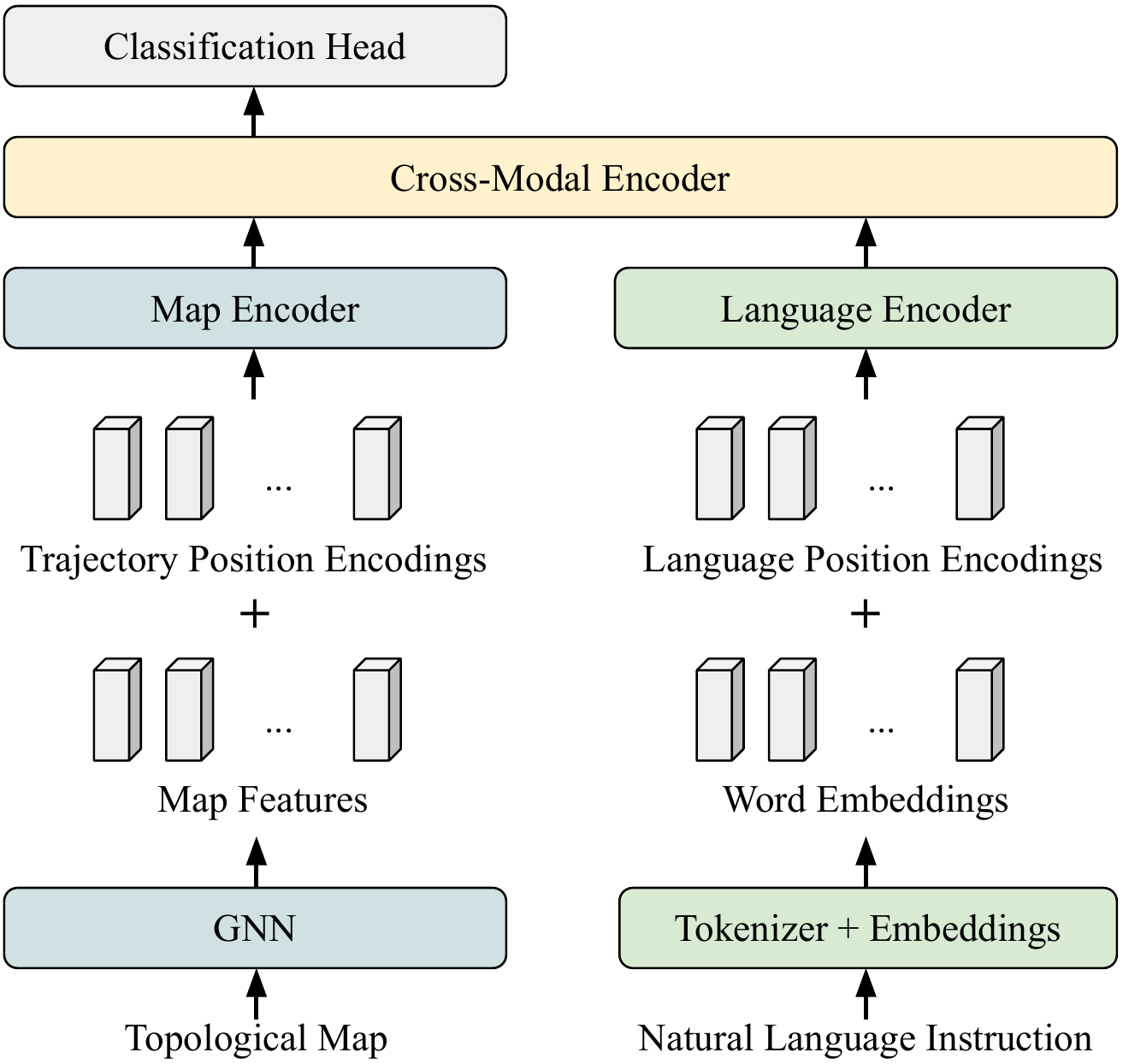}
\end{center}
\vspace{-1em}
\caption{\textbf{Planner.} The planner processes the topological map and language instruction separately. The information is fused with a cross-modal transformer (map, language, and cross-modal encoder) to classify the next step in the plan.}
\label{fig:planner-overview}
\vspace{-1em}
\end{figure}

As stated earlier, we modularize our approach into planning and control. For planning, the agent uses the constructed topological map and the navigation instruction to formulate a global navigation plan as a path in the map. This path, represented as a sequence of nodes, is then passed to the controller at the end of the planning stage.

As depicted in Fig. \ref{fig:planner-overview}, our planner has two main components: a graph neural network (GNN) and a cross-modal transformer (comprised of a map, language, and cross-modal encoder). The GNN computes representations of the environment which capture visual appearance and environment connectivity. These representations are passed along with the navigation instruction to the cross-modal transformer which selects the next node in the plan in a sequential, auto-regressive fashion. This process repeats until the planner classifies that the end of the plan has been reached.


\paragraph{Relationship to language modeling and translation.} Notably, our problem setup and approach bear similarities to language modeling and translation. In neural machine translation (NMT) specifically, the model conditions on the source sentence $x$ in addition to its prior predictions $y_1, \ldots, y_{t-1}$ in the predicted sentence. 
\begin{align}
    \label{eq:chain-rule}
    p(y|x) = \prod_{t = 1}^T p(y_t | y_1, \ldots, y_{t-1}, x)
\end{align}

Analogously, in our setup the agent predicts each step of the navigation plan $y_t$ conditioned on its previous predictions $y_1, \ldots, y_{t-1}$ as well as the topological map $G$ and the navigation instruction $L$ (i.e., $x = (G, L)$). This important insight allows us to approach the VLN planning problem in a manner similar to NMT and language modeling.

The following subsections illustrate the architectural setup of our approach (Fig.~\ref{fig:planner-overview}). The planner has two branches for encoding the map and the language instruction. In the map branch, we use a graph neural network (GNN) to process the topological map and generate map features (Sec.~\ref{sec:method-gnn}). In the language branch, the instruction is mapped to word embeddings. Finally, the map features and word embeddings are passed into the cross-modal transformer to produce a navigation plan (Sec.~\ref{sec:method-planner-architecture}).

\subsubsection{Learning Environment Representations using Graph Neural Networks (GNNs)}
\label{sec:method-gnn}

To facilitate learning a mapping between language and physical space, the map features passed to the transformer should encapsulate visual appearance and environment connectivity. In our work, these map features are learned via a graph neural network (GNN), which carries strong relational inductive biases appropriate for topological maps.

The GNN is composed of sequential graph network (GN) blocks \cite{battaglia2018relational}. Each GN block takes as input a graph $\tilde{G} = (\tilde{\mathbf{u}}, \tilde{V}, \tilde{E})$ and produces an updated graph $\tilde{G}' = (\tilde{\mathbf{u}}', \tilde{V}', \tilde{E}')$ where the output features are computed according to the graph structure. To do this, the GN block is comprised of update functions $\phi^v(\cdot)$, $\phi^e(\cdot)$, $\phi^u(\cdot)$ and aggregation functions $\rho^{e \rightarrow v}(\cdot)$, $\rho^{e \rightarrow u}(\cdot)$, $\rho^{v \rightarrow u}(\cdot)$. We implement the update functions as multi-layer perceptrons (MLP) and use summation for the aggregation functions. For more details on GNNs, we refer the reader to Battaglia \etal~\cite{battaglia2018relational}.

\paragraph{Input graph representation.} The input to the GNN is the topological map encoded as a directed graph $G = (\mathbf{u}, V, E)$. To capture visual appearance and semantics, we encode each vertex as a ResNet152 feature~\cite{he2016deep} extracted from the corresponding fixed orientation RGB panorama. To encode the relative geometry between nodes, we map each edge category (Sec.~\ref{sec:topomaps}) to a learnable embedding. Each edge thus captures information about the relative orientation and distance between a pair of connecting nodes. Lastly, the global feature $\mathbf{u}$ is also a learned embedding.

\paragraph{Output graph representation.} At the final layer of the GNN, we extract the output node features $\hat{V}=\{\mathbf{\hat{v}}_i\}_{i = 1:n}$ as the environment map features. Due to the message passing nature of the GNN, this set of map features not only captures visual appearance but also environment connectivity. The features are passed along with the tokenized instruction to the cross-modal transformer described in the next section.

\subsubsection{Cross-Modal Transformer}
\label{sec:method-planner-architecture}


At the core of our planner is a cross-modal transformer which takes as input the topological map encoded as map features (modality 1; Sec.~\ref{sec:method-gnn}) and the navigation instruction (modality 2). The start node is also provided. Each forward pass computes one step of the navigation plan. Our transformer is based on LXMERT~\cite{tan-bansal-2019-lxmert} which has been demonstrated to be effective for vision and language tasks.

\paragraph{Attention.} A fundamental component of this model is the self-attention layer \cite{vaswani2017attention}. Specifically, an input sequence $x \in \mathbb{R}^{n \times d}$ of length $n$ and dimension $d$ is linearly projected into a set of keys $K \in \mathbb{R}^{n \times d_k}$, queries $Q \in \mathbb{R}^{n \times d_k}$, and values $V \in \mathbb{R}^{n \times d_v}$. The queries and keys are then used to compute a weighted sum over the values.
\begin{align}
    \text{Attn}(Q, K, V) = \text{softmax} \Big( \frac{QK^\top}{\sqrt{d_k}} \Big) V
\end{align}

\paragraph{Single modal encoder.} Each modality (map and language) is encoded via its corresponding single-modality encoder. This encoder follows a standard transformer model comprised of self-attention layers and feedforward layers. As depicted in Fig.~\ref{fig:planner-overview}, the input to the map encoder is the sum of the map features (output from the GNN) with trajectory positional encodings. The language branch processes word embeddings from the natural language instruction summed with learned positional encodings.

\paragraph{Cross modal encoder.} The cross-modal encoder~\cite{tan-bansal-2019-lxmert} exchanges information between the two modalities. Each layer of the cross-modal encoder is composed of cross-modal attention, self-attention, and feed forward layers. In the cross-modal attention layer, the query comes from one modality, and the keys and values come from the other modality. The calculation for two input sequences $x_A, x_B$ corresponding with modalities $A$ and $B$ is in Eq.~\ref{eq:cross-modal-attention}. 
\begin{align}
    \label{eq:cross-modal-attention}
    \text{CrossAttn}(x_A, x_B) = \text{Attn}(Q_A, K_B, V_B)
\end{align}
The node features produced by the cross-modal encoder are passed to a classification head, implemented as an MLP, to compute the next step in the plan.

\begin{figure}
\begin{center}
   \includegraphics[width=0.95\linewidth]{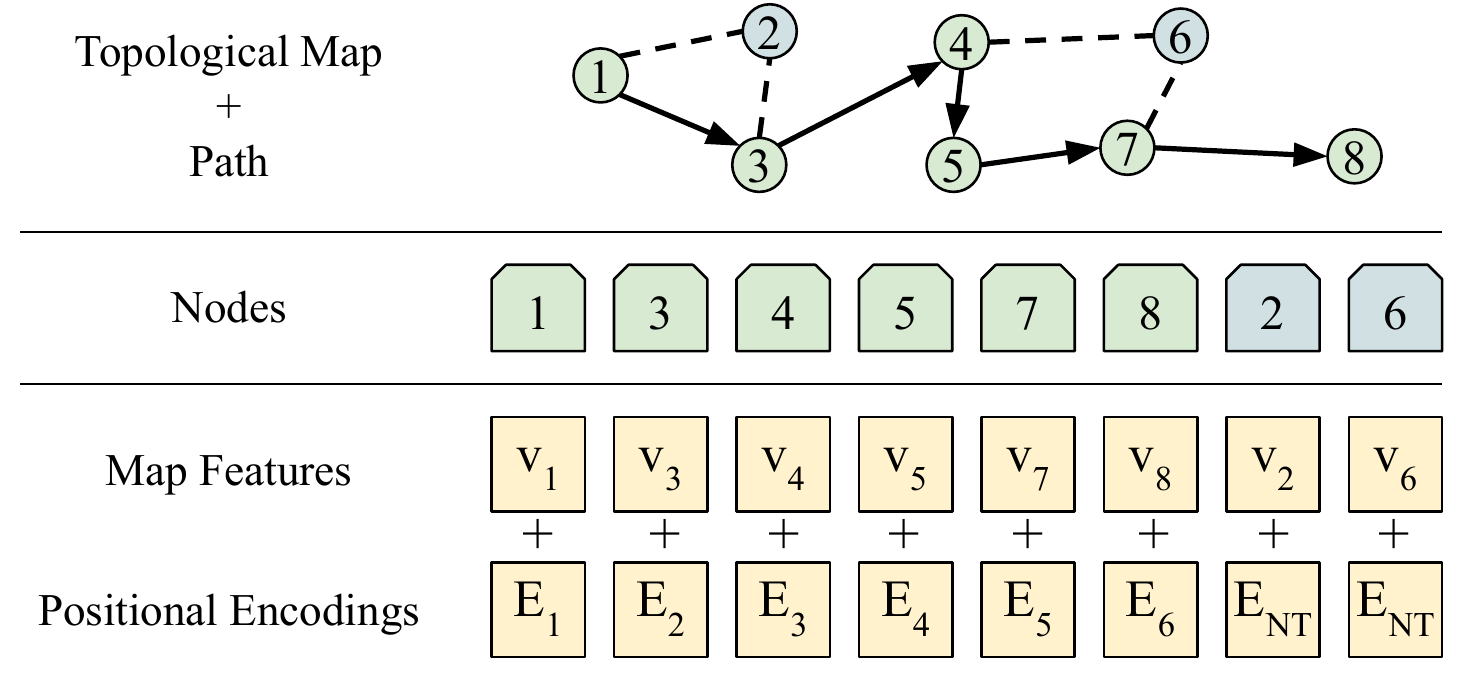}
\end{center}
\vspace{-1em}
   \caption{\textbf{Trajectory position encoding.} Nodes in the predicted trajectory (green) utilize a positional encoding for the corresponding position, whereas non-trajectory (NT) nodes (blue) utilize the same positional embedding $E_{NT}$.}
\label{fig:pos-encoding}
\end{figure}

\paragraph{Trajectory position encoding.} Since the transformer predicts a single node token at a time, it is essential that it keeps track of what has been predicted so far. To do this, we add positional embeddings to the node sequence as illustrated in Fig.~\ref{fig:pos-encoding}. While sinusoidal positional encodings are commonly used \cite{vaswani2017attention}, we found that learned positional embeddings yielded higher planning accuracy.

\paragraph{Stop action.} We use a \verb|[STOP]| token to indicate the stop action. It is appended to the map features and classification is performed over the nodes and the \verb|[STOP]| token. 

\paragraph{Training details.} The planner (GNN and transformer) is trained end-to-end with a cross entropy training objective. We use the AdamW optimizer \cite{kingma2014adam,loshchilov2018fixing} with a linear warmup schedule. Further architecture and implementation details are described in the appendix.


\subsection{Controller: Executing the Navigation Plan}
\label{sec:controller}

The controller is responsible for converting the plan (a topological path) into a series of low-level actions that take the agent to the goal. The inputs of the controller are the RGBD panorama of each planned node in the sequence, the current RGBD panoramic observation at each time step, and heading information. The output action space of the controller is defined as a set of parameterized low-level actions: {\sc forward} ($0.25m$), {\sc rotate left/right} (15\degree). 

The controller produces actions which move the agent from one node to another. To do this, it must also perform localization in order to determine when the current subgoal node has been reached, at which point the agent can move on to the next node in the plan.
We abstract our controller $C$ to two layers: a low level controller $C_{low}$ that moves the agent towards a geometric waypoint, and a high level controller $C_{high}$ that predicts such a waypoint. This high-level controller is used for localization.

\begin{figure}
\begin{center}
   \includegraphics[width=\linewidth]{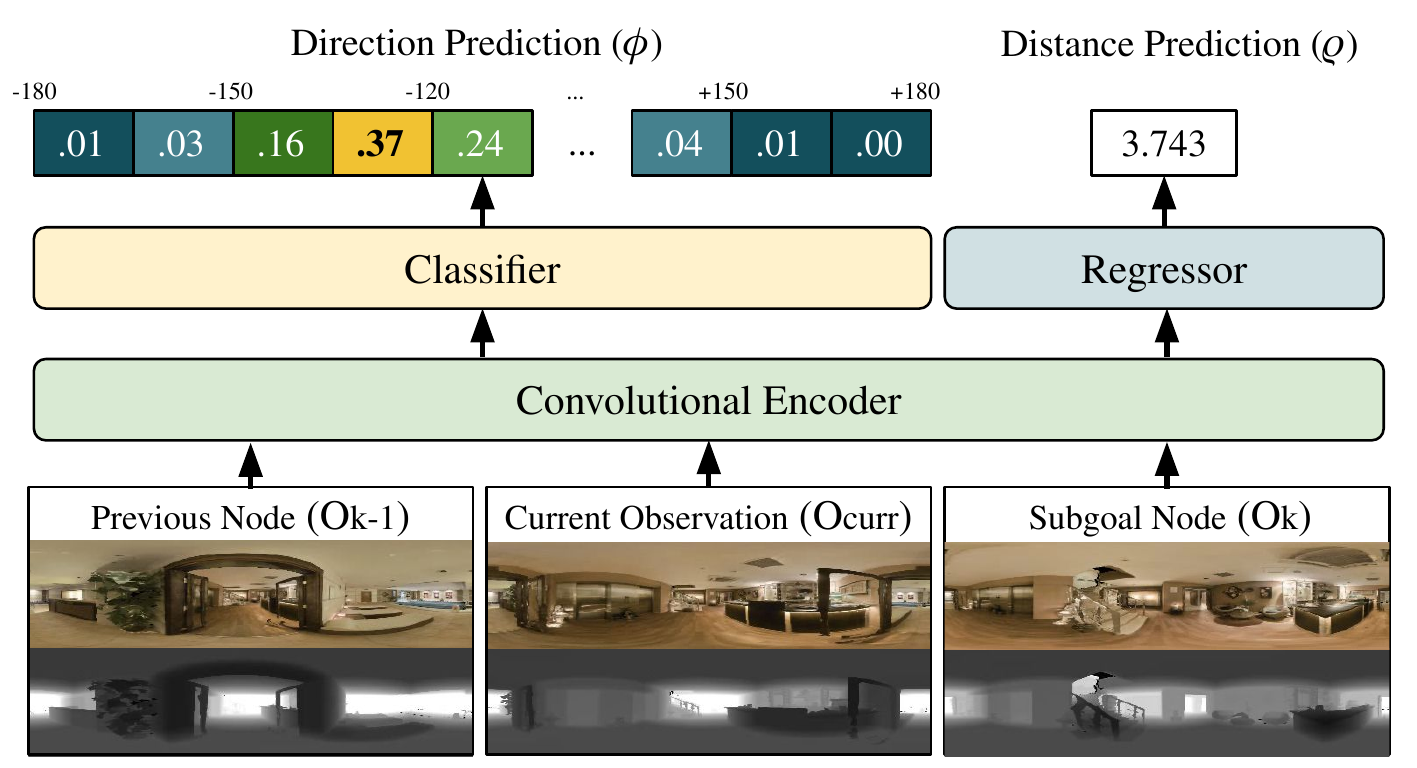}
\end{center}
\vspace{-1em}
   \caption{\textbf{High level controller}. $C_{high}$ uses the agent's observation and the previous and current subgoal node to predict a waypoint relative to the agent in polar coordinates. }
\label{fig:high-level-control}
\end{figure}

\paragraph{High level controller.} The high-level controller uses three panoramas, including the current and subgoal observations, to predict a waypoint to move towards in order to reach the subgoal. We use a quantized polar coordinate system to describe the waypoint relative to the agent. The polar coordinate is partitioned to 24 sectors, encompassing 15 degrees each. A sector index and a numeric distance defines the waypoint position. The predicted distance is used for determining whether the subgoal has been reached (localization).


Figure \ref{fig:high-level-control} illustrates the design of the high level controller $C_{high}$. The partially observable agent state at any time step is represented by a tuple of RGBD panoramas $(o_{k-1}, o_{curr}, o_{k})$ where $k$ is the index of the current subgoal. The observations correspond with the previous subgoal node, the current position, and the current subgoal node, respectively. All observations are oriented in the same heading direction using the heading information. To optimize for panoramic observations, we use circular convolution layers \cite{schubert2019panoconv} to encode the observations into features. The features are then used to predict a classification vector $\phi$ (direction of the subgoal node) and a scalar $\rho$ (distance between the agent and the subgoal node).

Both outputs are optimized simultaneously by imitation learning using dataset aggregation (DAgger) \cite{ross2011reduction}, supervised by A* expert waypoints of a fixed lookahead distance. The controller is trained separately from the planner.

\paragraph{Handling Oscillation.} Because the agent's state is defined by three panoramic observations only, there exists the possibility of perceptual aliasing in which the agent's observation is similar in multiple directions (e.g. a symmetric-looking hallway with an anchor node at each end). In this case, it is possible for the agent to move towards either direction in alternating time steps, resulting in no progress.

To alleviate this issue, we introduce ``stubbornness'' to the predicted heading probabilities such that the predicted direction is biased towards the prediction of the previous time step.
We define a bias function $B(\phi'_{t-1}, \phi_{t}; \sigma^2)$ to reweight the polar prediction at time $t$ by multiplying each element by a Gaussian distribution centered at the predicted direction at the previous time step:
\begin{align}
    \phi'_{t_i} = \phi_{t_i} * N(i-\arg\max_j(\phi'_{{t-1}_j}), \sigma^2)
\end{align}
where $i,j$ are indices into $\phi$, and $\sigma^2$ is the variance of the Gaussian such that the lower the variance, the more biased the waypoint is towards the previous predicted direction.

\paragraph{Mapping to low-level actions.} After $C_{high}$ produces a waypoint, a robot-specific low-level controller translates it to a robot action. For our problem setup, we define $C_{low}$ as a simple policy that maps $\phi'_t$ to quantized low-level actions: {\sc forward} if agent is facing the direction of $\arg\max_i \phi'_{t_i}$; otherwise {\sc rotate} towards that direction. This can be replaced by more sophisticated robot-specific controllers.

\paragraph{Localization and trajectory following.} When the agent traverses within a threshold distance $d$ to the current subgoal node such that $\rho \le d$, it consumes that node. The subsequent node in the plan becomes the new subgoal. This process is repeated until no more nodes remain in the plan.

\section{Experiments}
\label{sec:experiments}

We evaluate our approach using the Interactive Gibson simulator (iGibson) \cite{xia2019gibson,xia2020interactive}. The agent is equipped with a $360\degree$ panoramic RGBD camera and ground truth heading information. Prior to navigation, the agent explores each environment via 10 pre-defined trajectories and constructs topological maps denoted as agent-generated (AG) graphs. For evaluation, we use the VLN-CE dataset \cite{krantz2020navgraph} and evaluate on environments which were seen (val-seen) and unseen (val-unseen) during training. We use the following metrics: success rate (SR), oracle success rate (OS), and navigation error from goal in meters (NE). Oracle success rate uses an oracle to determine when to stop. A navigation episode is successful if the agent stops within $3m$ of the goal position. Details on these metrics can be found in \cite{anderson2018evaluation,anderson2018vision}.


\subsection{Planner Evaluation}
\label{sec:experiments-planner}

In this section, we evaluate the performance of the planner in isolation from the controller. In addition to the agent-generated maps, we also compare planning performance on the pre-defined navigation graphs from Room2Room (R2R) \cite{anderson2018vision}. We compare our approach with a graph neural network (GNN) baseline as well as variants of our cross-modal transformer planner (CMTP) with different stop mechanisms:

\begin{itemize}[align=left,labelsep=0.3em,leftmargin=*,itemsep=0em,parsep=0.3em]
    \item[--] \textbf{Graph neural network (GNN)}: A GNN that directly predicts the final navigation goal in the topological map when given the map, start node, and instruction. The instruction is encoded using a transformer and passed as the global feature to the GNN. The GNN then performs classification over the nodes. To extract a navigation plan, we compute a shortest path from the start node to the predicted goal. This baseline is similar in spirit to \cite{deng2020evolving}. 
    \item[--] \textbf{Cross-Modal Transformer Planner (CMTP)}: Our planner which predicts the stop action by performing classification over the \verb|[STOP]| token and node tokens.
    \item[--] \textbf{CMTP Binary Cross Entropy (CMTP-BCE)}: Uses a separate (binary) cross-entropy loss for the \verb|[STOP]| token, trained jointly with the node classification loss.
    \item[--] \textbf{CMTP Repeated (CMTP-R)}: Instead of using a \verb|[STOP]| token, this CMTP is trained to indicate the stop action by predicting the same node consecutively.
    \item[--] \textbf{CMTP No Stop (CMTP-NS)}: Never produces a stop command and is instead used as an ablation. It provides insight into planning performance when correctly timing the stop action is taken out of the picture. The episode terminates after 20 planning steps.
\end{itemize}

\begin{table}[t]
    \centering
    \small
    \setlength{\tabcolsep}{4pt}
    \begin{tabular}{clcccccc}
        \toprule
        & & \multicolumn{3}{c}{Val-Seen} & \multicolumn{3}{c}{Val-Unseen} \\
        \cmidrule(l){3-5} \cmidrule(l){6-8}
        & & SR $\uparrow$ & OS $\uparrow$ & NE $\downarrow$ & SR $\uparrow$ & OS $\uparrow$ & NE $\downarrow$ \\
        \midrule
        & GNN & 30.6 & 45.6 & 8.04 & 23.5 & 40.7 & 8.15 \\
        & CMTP & \textbf{37.9} & \textbf{54.0} & 6.54 & 26.0 & 39.1 & \textbf{7.62} \\
        AG & CMTP-BCE & 33.6 & 48.1 & 7.11 & \textbf{27.3} & \textbf{42.9} & 8.13  \\
        & CMTP-R & 37.4 & 52.6 & \textbf{6.28} & 26.5 & 41.3 & 7.66 \\ \rowcolor{Gainsboro!60}
        & CMTP-NS & - & 66.2 & - & - & 64.0 & - \\
        \midrule
        & GNN & 27.3 & 51.2 & 9.62 & 20.7 & \textbf{49.5} & 11.21 \\
        & CMTP & \textbf{38.6} & 49.1 & \textbf{6.76} & 28.7 & 37.6 & \textbf{7.74} \\
        R2R & CMTP-BCE & 37.1 & \textbf{51.6} & 6.97 & \textbf{30.2} & 42.2 & 8.36 \\
        & CMTP-R & 38.4 & 49.8 & 6.91 & 26.8 & 36.7 & 8.04 \\ \rowcolor{Gainsboro!60}
        & CMTP-NS & - & 61.4 & - & - & 60.5 & - \\
        \bottomrule
    \end{tabular}
    \caption{\textbf{Planner performance.} Our CMTP models outperform the baseline GNN and perform better on R2R graphs. CMTP-NS is provided as an ablation. See text for details.}
    \label{tab:quantitative-results-planner}
\end{table}

Quantitative results are presented in Table~\ref{tab:quantitative-results-planner}.
The transformer planner achieves higher success rates and lower navigation errors compared to the GNN. Across all models, the performance on seen environments is significantly higher than unseen environments. This overfitting is most prominent in the CMTP model. On the other hand, the CMTP-BCE model has less overfitting at the cost of lower performance on seen environments.

We observe that the oracle success rate of CMTP-NS is significantly higher than all models. In other words, the transformer planner often navigates to the correct locations but struggles to stop at the right time. These results highlight the challenge of accurately stopping at the right moment while illustrating the capability of our model for choosing the correct places to navigate towards.
Different topological maps also affect planning performance. For example, the transformer planners achieve higher performance on the pre-defined R2R graphs. This may be due to better positioned nodes and edges compared to the AG maps, which sometimes has awkwardly placed nodes near walls or corners. Importantly, these results indicate the importance of (1) developing models which are able to navigate consistently well under different topological maps and (2) exploring new memory representations which support effective and robust navigation.

\begin{figure}
\begin{center}
   \includegraphics[width=\linewidth]{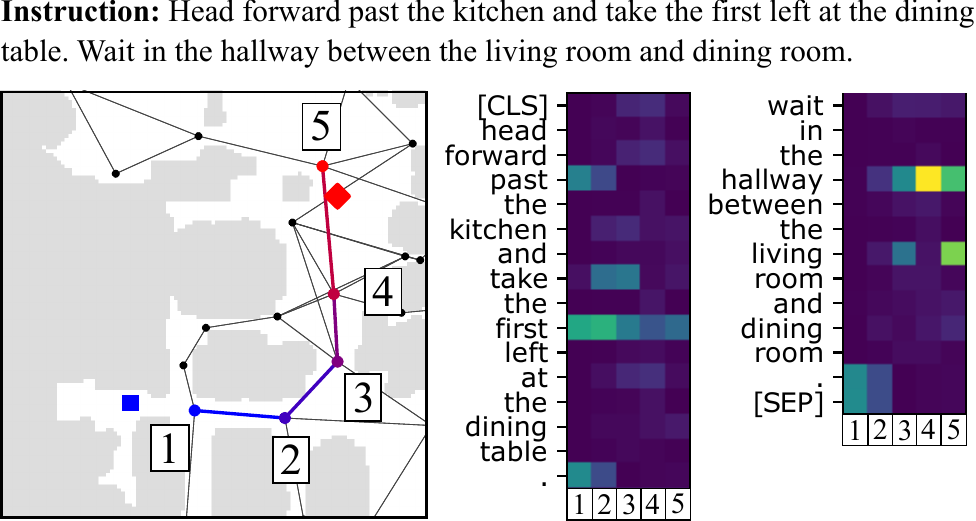}
\end{center}
\vspace{-1em}
  \caption{\textbf{Cross modal attention} (right) from nodes onto instructions (top). Each column represents a node in the predicted plan (left), going from blue to red.} 
\label{fig:xmodal-attention}
\vspace{-1em}
\end{figure}

\paragraph{Cross modal attention.} The cross modal attention design of our planner facilitates informational flow between the topological map and language domains, which is intended to ground correlated visual node features to their instructional counterparts. An example is shown in Fig \ref{fig:xmodal-attention}, which visualizes the first layer cross modal attention from predicted nodes to the input instruction. We see that words which encode the first half of the instruction correspond with earlier nodes in the trajectory, whereas words such as ``hallway'' and ``living room'' in the latter half of the sentence correspond to the last three nodes of the trajectory.

We note that the model will sometimes focus attention on articles or punctuation. This behavior is in line with previous observations \cite{tan-bansal-2019-lxmert} that the cross modal layer tends to focus on articles over single noun references (e.g. attention on ``the'' or ``living'' instead of ``room''), using them as anchor words while likely relying on the language encoder to connect the articles to the specific objects they reference.

\subsection{Controller Evaluation}
\label{sec:experiments-controller}

\begin{figure}
\begin{center}
   \includegraphics[width=1.0\linewidth]{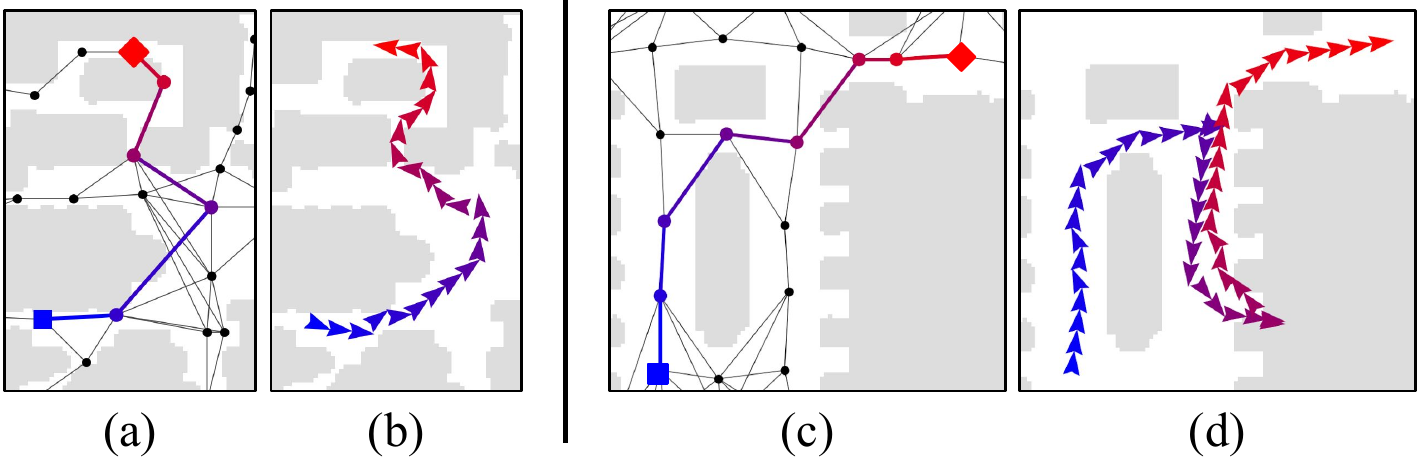}
\end{center}
\vspace{-1em}
\caption{\textbf{Plan execution}. The controller is able to follow the navigation plans (a, c) as shown by the trajectories (b, d). It can also correct for mistakes via backtracking (d).}
\label{fig:controller-mistake}
\end{figure}

\begin{table}[]
\centering
\small
\setlength{\tabcolsep}{7pt}
\begin{tabular}{@{}ccccc@{}}
\toprule
& 
Unbiased & $\sigma^2 = 5$ & $\sigma^2 = 3$ & $\sigma^2 = 2$ 
\\ \midrule
Pointgoal &   
99.8 / 99.5 & 99.9 / 99.4 & 99.9 / 99.7 & 99.6 / 99.2
\\ \midrule
AG  &  
83.3 / 82.5 & 89.4 / 87.2 & 93.1 / 89.6 & 77.7 / 75.9       
\\ 
R2R  &  
87.2 / 82.0 & 93.3 / 89.1 & 92.8 / 92.6 & 84.3 / 81.8
\\ \bottomrule
\end{tabular}
\caption{\textbf{Controller success rate} in [seen / unseen] environments for random point goals ($\le 5m$ away) and for randomly sampled plans from R2R or AG graphs ($\le 12$ nodes away). Adding the bias greatly improves performance.}
\label{tab:controller-success}
\vspace{-1em}
\end{table}

\begin{table*}[]
\centering
\small
\setlength{\tabcolsep}{7pt}
\begin{tabular}{@{}clcccccccccccc@{}}
\toprule
&
& \multicolumn{6}{c}{Val-seen} & \multicolumn{6}{c}{Val-unseen} \\ 
  \cmidrule(l){3-8} \cmidrule(l){9-14}
&
& PS$\uparrow$ & CS$\uparrow$ & SR$\uparrow$ & OS$\uparrow$ & NE$\downarrow$ & SPL$\uparrow$ 
& PS$\uparrow$ & CS$\uparrow$ & SR$\uparrow$ & OS$\uparrow$ & NE$\downarrow$ & SPL$\uparrow$ \\
\midrule
&
VLN-CE \cite{krantz2020navgraph}
& -  & -  & 22.2  & 29.8  & 7.6 & 16.5     
& -  & -  & 19.7  & 27.2  & \textbf{7.8} & 14.1    \\
\midrule
\multirow{4}{*}{AG}
& 
GNN
  & 30.6  & 91.7  & 29.6  & 46.0  & 8.2  & 26.3
  & 23.5  & 84.8  & 22.1  & 40.0  & 8.4  & 17.2
  \\
& 
CMTP 
  & 37.9  & 92.2  & 35.9 & \textbf{56.2}  & 6.6  & 30.5
  & 26.0  & 86.6  & 23.1  & 39.2  & 7.9  & 19.1
\\ 
& 
CMTP-BCE 
  & 33.6  & 89.4  & 32.6  & 49.6  & 7.2  & 27.5
  & 27.3  & 82.6  & 25.2  & 42.2  & 8.4  & 20.2
\\
&
CMTP-R  
  & 37.4  & 93.1  & \textbf{36.3}  & 53.9  & \textbf{6.5}  & \textbf{31.3}
  & 26.5  & 87.8  & 25.3  & 42.6  & 7.9  & 20.3
\\
\midrule

\multirow{4}{*}{R2R}
&
GNN
  & 27.3  & 89.1  & 26.6  & 51.9  & 9.4  & 21.2
  & 20.7  & 88.6  & 20.3  & \textbf{48.3}  & 10.9  & 16.4
\\
&
CMTP 
  & 38.6  & 91.7  & 36.1  & 45.4  & 7.1  & 31.2
  & 28.7  & 90.0  & 26.4  & 38.0  & 7.9  & 22.7
\\ 
&
CMTP-BCE 
  & 37.1  & 89.6  & 34.4  & 49.3  & 7.4  & 29.9
  & 30.2  & 89.6  & \textbf{28.9}  & 40.7  & 8.4  & \textbf{24.1}
\\
&
CMTP-R 
  & 38.4  & 89.6  & 35.1  & 48.2  & 7.5  & 31.2
  & 26.8  & 90.0  & 25.2  & 36.0  & 8.3  & 20.9
\\
\bottomrule
\end{tabular}
\caption{\textbf{Integrated system performance.} The modular approaches (GNN, CMTP) outperform the end-to-end VLN-CE model in success rate (SR), with our CMTP models achieving the best performance on seen and unseen environments.}
\label{tab:integration}
\vspace{-1em}
\end{table*}

We evaluate the controller separate from our planner by executing episodes of random plans in the simulated environment. A navigation success threshold distance of $1m$ is used when evaluating the controller alone.

Fig.~\ref{fig:controller-mistake} shows qualitative results of the controller executing navigation plans. The controller is able to navigate while avoiding collision, despite graph edges cutting through obstacles in the planned path. With the inclusion of the previous node into the controller input, the agent is able to learn intelligent backtracking behavior in which the agent backtracks to the previously known position when it makes a mistake and deviates from the path.

The quantitative results reported in Table \ref{tab:controller-success} support our observations that the controller works well. In navigation to point goals (row 1), it achieves high success rates even in unseen environments. Executing multi-node plans (rows 2-3) also produces high success rates, with better performance on R2R graphs than AG graphs due to their better positioned nodes providing more informative observations.

\subsection{VLN Evaluation}
\label{sec:experiments-full-vln}

Finally, we evaluate our full navigation system. We compare with the cross-modal approach in VLN-CE \cite{krantz2020navgraph} since it is the only prior work, to our knowledge, that considers low-level actions like our work. VLN-CE is trained end-to-end and utilizes a gated recurrent unit (GRU)~\cite{cho2014learning} to encode the agent's history. In addition to prior metrics, we report success weighted by path length (SPL) \cite{anderson2018evaluation}, and planner success (PS) and controller success (CS) for modular approaches.

Table~\ref{tab:integration} shows that our CMTP models achieve the highest VLN success rates and SPL compared to the GNN and VLN-CE baselines. Moreover, the methods which use topological maps outperform the end-to-end VLN-CE baseline. While these results may be expected given that the former methods pre-explore the environments, the gap in performance is also due to reducing the search space through the topological maps. By combining our cross-modal planner and controller into an integrated system, we get the best of both worlds: the planner achieves high planning success rates while the controller is able to bring those plans to fruition with low-level discrete actions.

\begin{figure}
\begin{center}
   \includegraphics[width=1.0\linewidth]{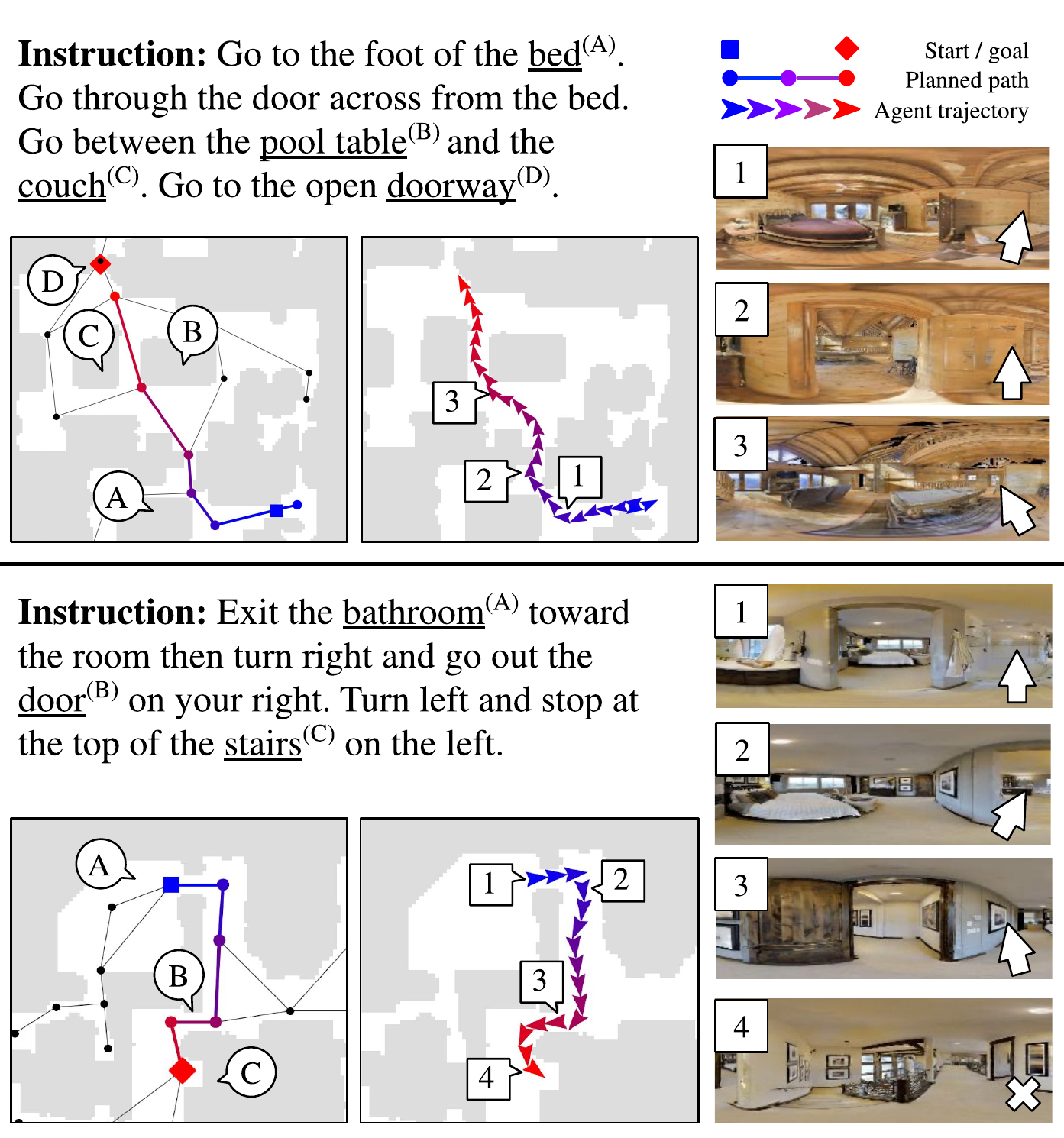}
\end{center}
\vspace{-1em}
\caption{\textbf{Our full navigation system.} We show the navigation instruction (top-left), the panorama observations (right), and the predicted navigation plan (bottom-left) and execution (bottom-middle). The arrows in each panorama indicate the predicted action at the given panorama.}
\label{fig:integrated-system}
\vspace{-2em}
\end{figure}

Fig.~\ref{fig:integrated-system} shows exciting qualitative results of our full navigation system. As illustrated by the agent's predicted plan and executed trajectory, the modular approach makes it easy to interpret the agent's intents and evaluate its ability to realize those intents. For instance, in Fig.~\ref{fig:integrated-system} (top), the planner predicts a path between the pool table and couch as specified by the instruction. The controller successfully executes the plan by converting it into a smooth trajectory which traverses between the two objects rather than around them.

\section{Conclusion}

We introduced a novel approach for VLN with topological maps. By using a modular system in conjunction with topological maps, we saw a boost in navigation performance compared to an end-to-end baseline. Potential extensions include demonstrating VLN on a real robot~\cite{anderson2020sim} and using topological maps constructed at test time \cite{chaplot2020neural}. We hope our work inspires further research in building robot systems which communicate effectively and operate robustly.

\paragraph{Acknowledgments.} This work was funded through the support of ONR MURI Award \#W911NF-15-1-0479. We also thank Stanford Vision and Learning Lab members for their constructive feedback, including Eric Chengshu Li and Claudia Pérez D'Arpino.

{\small
\bibliographystyle{ieee_fullname}
\bibliography{egbib}
}

\newpage

\appendix
\section{Topological Maps}

\subsection{Exploration Trajectories}

As a first step prior to navigation, the agent builds an understanding of the environment in the form of a topological map. To do this, it uses a set of exploration trajectories. Methods for exploring environments have been investigated in prior work (e.g. \cite{chaplot2019learning,chen2019learning}). In our work, we generate the exploration trajectories by sampling a set of waypoints in the traversable areas of the environments using the ground truth metric maps and then having the agent navigate to each of them. This allows us to create a fixed set of exploration trajectories to make the experiments and evaluation across baselines consistent. During the exploration process the agent keeps track of its observation at each time step as well as the odometry sensor reading of its position and orientation such that the agent is aware of the relative position between any two observations. The result is a dense trajectory throughout the environment as visualized in Fig.~\ref{fig:expl-traj-examples}.

\subsection{Map Construction}


\subsubsection{Reachability Estimator}

\begin{figure*}
\begin{center}
  \includegraphics[width=0.9\linewidth]{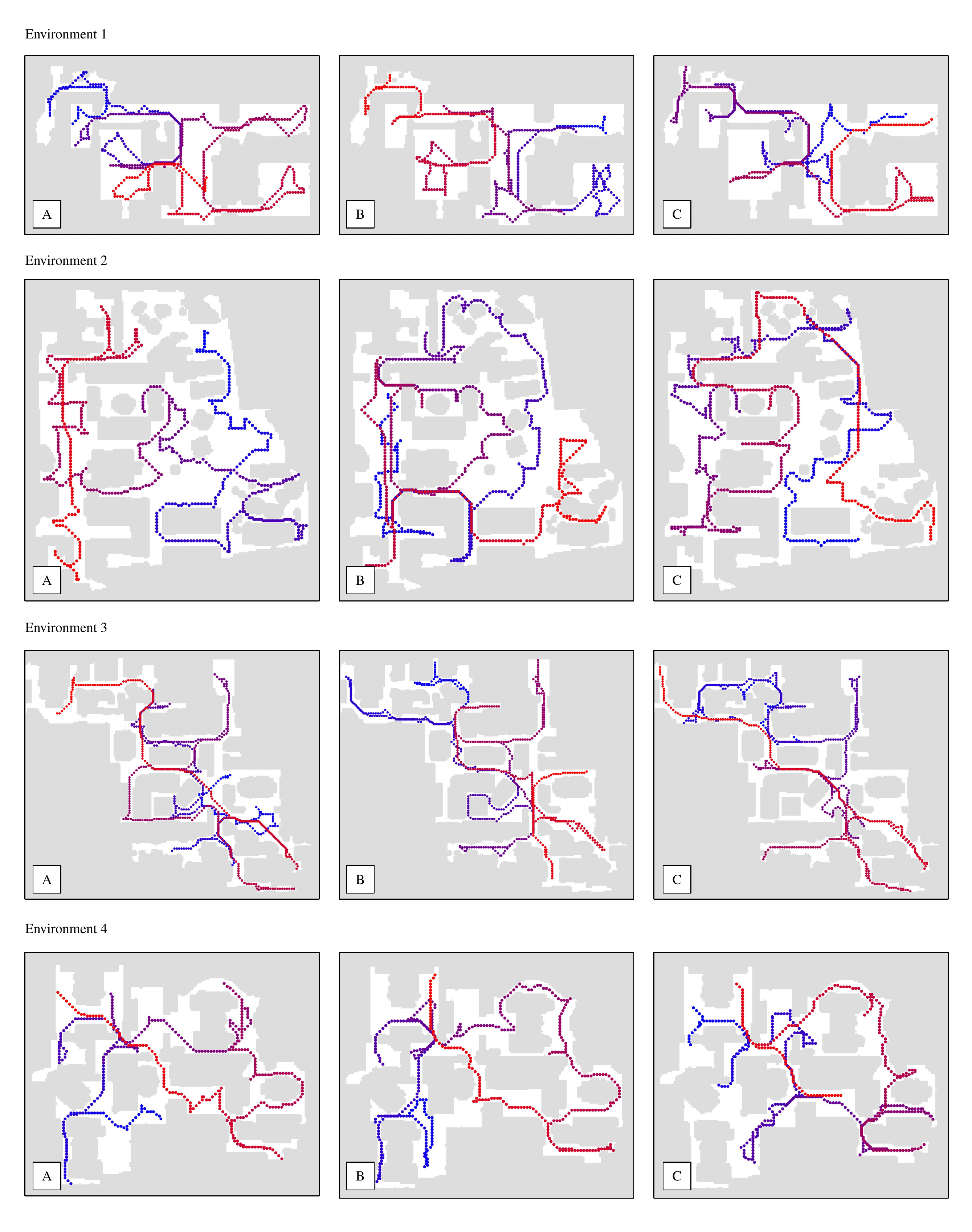}
\end{center}
  \caption{\textbf{Exploration trajectories.} Each row shows different exploration trajectories, going from blue to red, for a single environment. White represents traversable regions, while gray represents untraversable areas.}
\label{fig:expl-traj-examples}
\end{figure*}

\begin{figure*}
\begin{center}
  \includegraphics[width=0.9\linewidth]{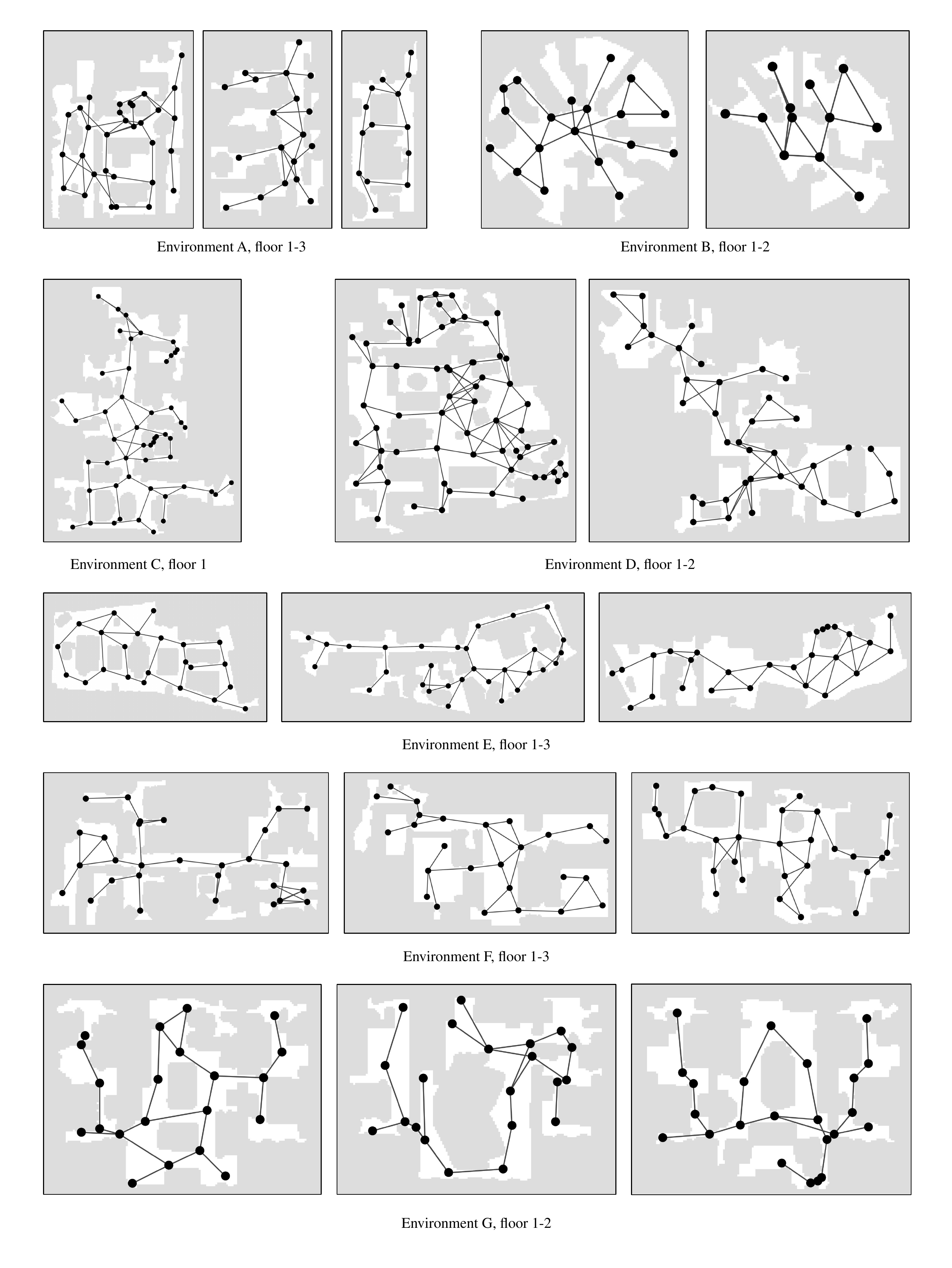}
\end{center}
  \vspace{-2em}
  \caption{\textbf{Agent-generated topological maps.} The graphs are reasonably sparse but still cover a large portion of the traversable space. Certain nodes may be difficult to reach or close to walls, such as in Environment B floor 1.}
\label{fig:ag-topo}
\end{figure*}

Examples of our agent-generated topological maps are shown in Fig.~\ref{fig:ag-topo}. We use a reachability estimator~\cite{meng2019scaling} to sparsify the exploration trajectories into a topological map. A reachability estimator is a function $RE$ that predicts the probability of any given controller to start from a certain position and reach a certain target. In our work, we use an oracle reachability estimator $RE$ that exploits the traversability of the environment. Here, we define reachability between two positions $p_1$ and $p_2$ as:
\begin{equation}
\begin{split}
    & RE(p_1, p_2) = \delta(p_1, p_2) * \max(\frac{d_{sp} - \|p_1 - p_2\| }{d_{sp}}, 0)
\end{split}
\end{equation}
where $\delta(p_1, p_2)$ represents line-of-sight between positions $p_1$ and $p_2$ according to the traversability maps (as shown in white and gray in Fig.~\ref{fig:expl-traj-examples}) and $d_{sp}$ is a parameter representing the sparsity of the resulting topological maps. In other words, reachability is nonzero if an unobstructed line can be drawn between $p_1$ and $p_2$ and if they are within $d_{sp}$ from each other.

\begin{figure*}
\begin{center}
  \includegraphics[width=0.9\linewidth]{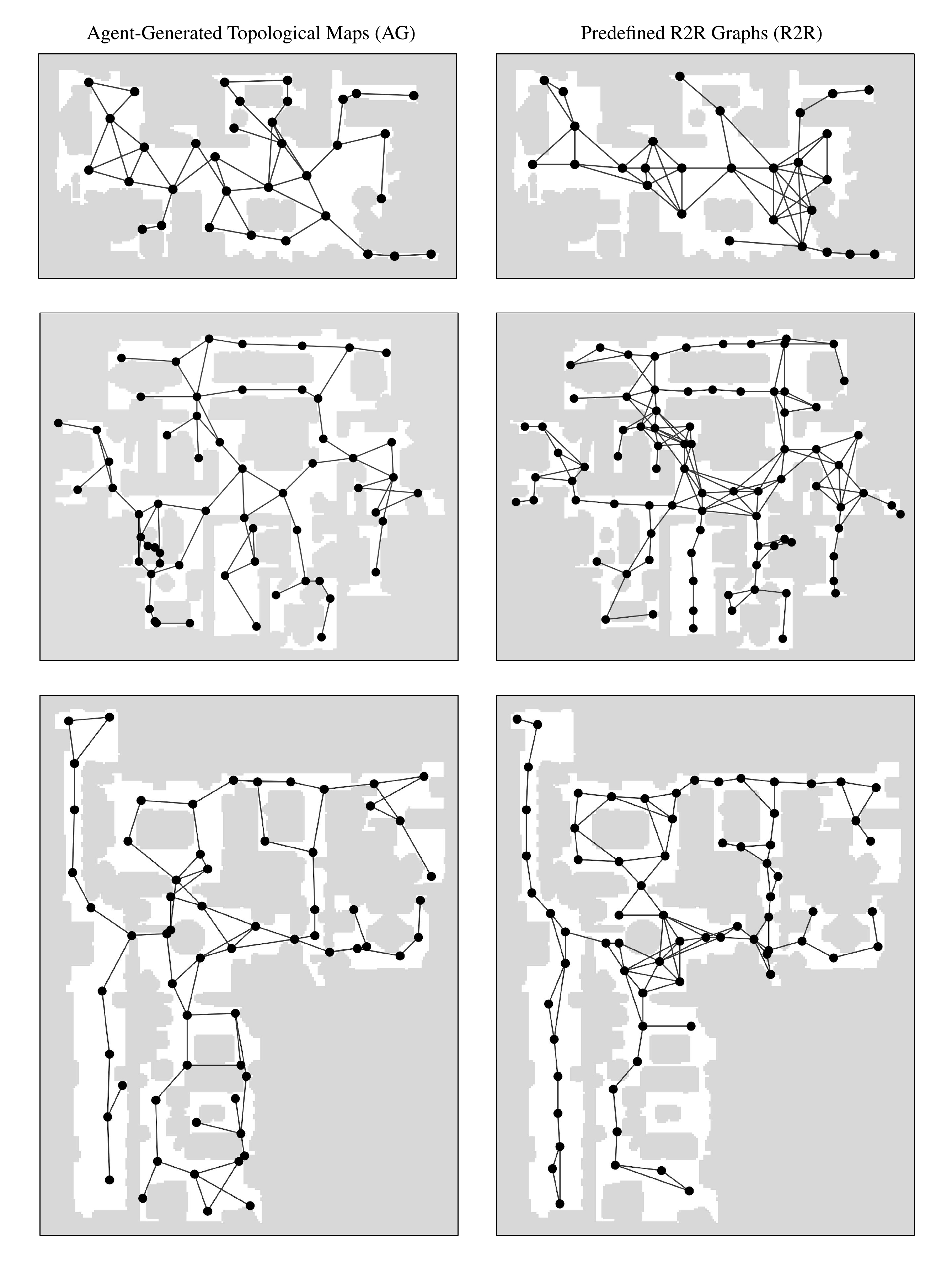}
\end{center}
  \caption{\textbf{Comparison of agent-generated maps (left) and R2R predefined graphs (right).} In general, the agent-generated maps are sparser, whereas the R2R graphs may have dense clusters of nodes.}
\label{fig:topo-compare}
\end{figure*}

\begin{figure}
\begin{center}
  \includegraphics[width=0.9\linewidth]{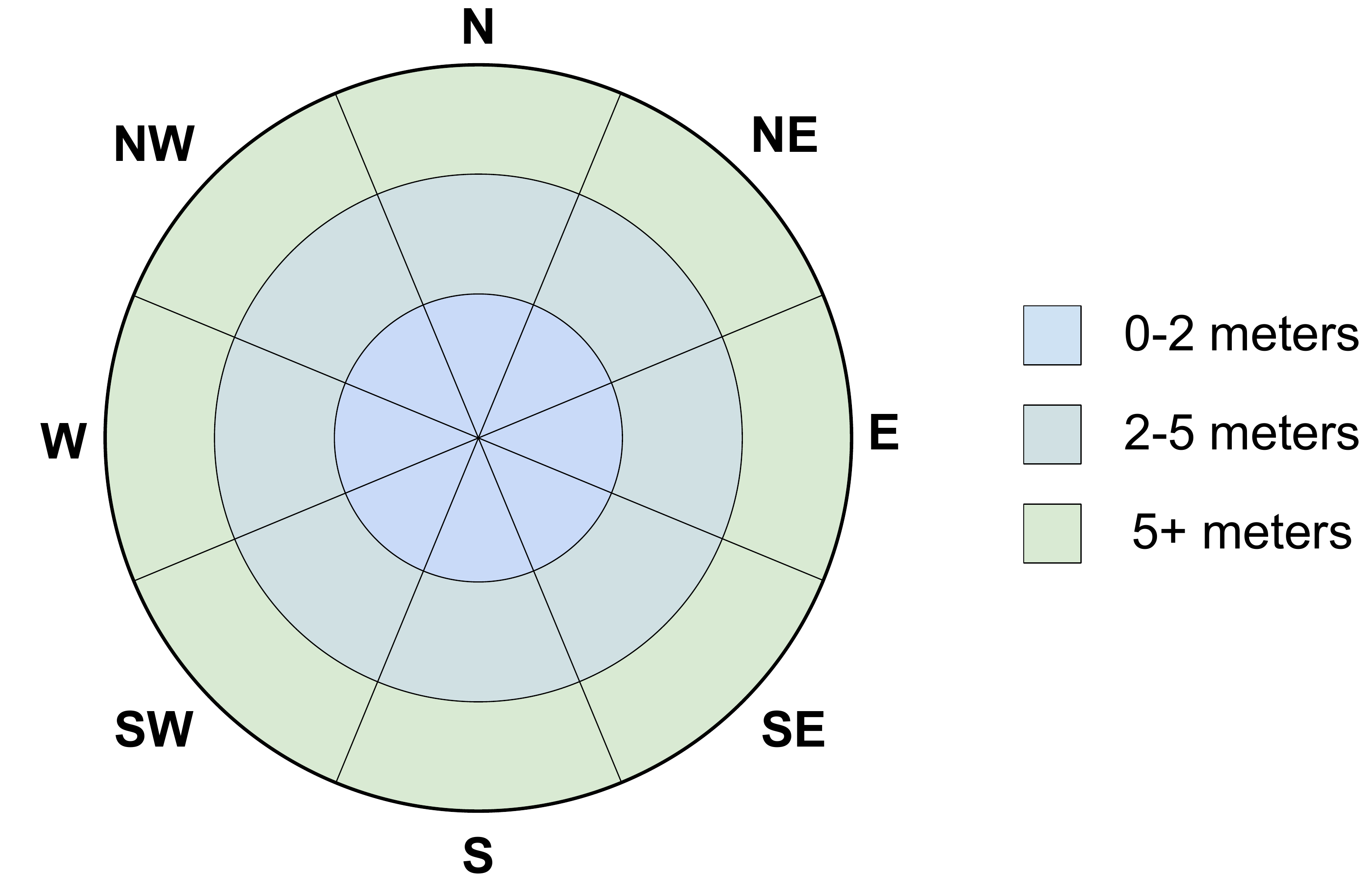}
\end{center}
  \caption{Directed edge labels for the topological map are represented as discretized polar coordinates. The chart is represented as a top-down view with the agent in the center, and orientations represented as compass directions.}
\label{fig:polar-coord}
\end{figure}

\subsubsection{Sparsifying a dense trajectory}

Given a dense exploration trajectory, for any starting position $p_i$, we may confidently discard any position $p_{i+1}...p_{j-1}$ as long as $RE(p_i, p_j)$ is sufficiently high \cite{meng2019scaling}. In other words, we greedily choose the next node by
\begin{equation}
\begin{split}
    & \max j \\
    & \text{s.t. } RE(p_i, p_k) > p_{sparse}, \forall k, i<k\le j
\end{split}
\end{equation}
where $p_{sparse}$ is the threshold that guarantees a minimum reachability of any edge in the resultant graph. 

Each exploration trajectory produces one sparsified topological map. Since we perform multiple exploration runs per environment, we merge multiple topological maps into one by adding edges between any two nodes $i$ and $j$ similarly if $ RE(p_i, p_j) > p_{sparse}$. 

Finally, we further sparsify the resultant singular graph by merging nodes $i$ and $j$ if $ RE(p_i, p_j) > p_{merge}$, where $p_{merge}$ is a threshold above which two nodes are merged into a single one. All neighbors of $i$ and $j$ become the neighbors of the new node, except the ones with reachability dropping below $p_{sparse}$ after the merge. 

In our setup, we use $d_{sp} = 4m$, $p_{sparse} = 0$, $p_{merge} = 0.5$, such that edges are established between positions reachable in a straight line within $4m$ of each other, and nodes within $2m$ of each other are merged. 

\subsubsection{Comparison to Pre-Defined R2R Graphs}

We illustrate differences between our agent-generated (AG) maps and the predefined Room2Room (R2R) graphs~\cite{anderson2018vision} in Fig.~\ref{fig:topo-compare}. The AG maps are sparser and may have nodes which are very close to the walls. For instance, as shown in the top row of Fig.~\ref{fig:topo-compare}, in the R2R graph the hallway along the center of the map has nicely positioned nodes. However, the nodes in the AG maps for the same space are positioned along the walls, resulting in slightly more redundant nodes.

We also see large node clusters in open spaces in the R2R graphs. In these locations, there are several nearby nodes in the same area as well as edges which connect the nodes with each other. This can be seen in every R2R graph in Fig.~\ref{fig:topo-compare}.

\subsection{Topological Map Representation}

As stated in Sec.~\ref{sec:topomaps}, each topological map node is encoded as a ResNet152 feature~\cite{he2016deep}, and each edge is mapped to an embedding corresponding with the edge category. The edge categories are in discretized polar coordinates which are visualized in Fig.~\ref{fig:polar-coord}.

\section{Method}

\subsection{Cross-Modal Planning}

\subsubsection{Graph Neural Network (GNN)}

Our graph neural network consists of 6 sequential graph network (GN) blocks. Borrowing the notation from Battaglia \etal~\cite{battaglia2018relational}, each GN block is comprised of update functions $\phi^v(\cdot)$, $\phi^e(\cdot)$, $\phi^u(\cdot)$ and aggregation functions $\rho^{e \rightarrow v}(\cdot)$, $\rho^{e \rightarrow u}(\cdot)$, $\rho^{v \rightarrow u}(\cdot)$.

\paragraph{Update functions.} We implement all update functions $\phi^{(\cdot)}$ as multi-layer perceptrons (MLPs). Each MLP has 4 fully-connected layers of hidden dimension 1024, each followed by batch normalization~\cite{ioffe2015batch} and ReLU except the last layer.

\paragraph{Aggregation functions.} We implement all aggregation functions $\rho^{(\cdot)}$ as element-wise summations.

\paragraph{Inputs to first GN block.} The first GN block the same dimensions as the intermediate GN blocks. As input to the first GN block, we use the following:
\begin{itemize}
    \item Node features $\mathbf{v}_i^{(1)} \in \mathbb{R}^{1024}$ for the $i$th node in the topological map. Each node embedding is a ResNet152 embedding. 
    \item Edge features $\mathbf{e}_j^{(1)} \in \mathbb{R}^{256}$ for the $j$th edge in the topological map. Each edge embedding is a learned feature for the corresponding edge category (Fig.~\ref{fig:polar-coord}). 
    \item A single global feature $\mathbf{u}^{(1)} \in \mathbb{R}^{512}$ which is a learned embedding.
\end{itemize}

\paragraph{Intermediate GN blocks.} The intermediate GN blocks use the following input and output dimensions for intermediate layer $k$:
\begin{itemize}
    \item Node features $\mathbf{v}^{(k)}_i \in \mathbb{R}^{1024}$ for the $i$th node in the topological map.
    \item Edge features $\mathbf{e}^{(k)}_j \in \mathbb{R}^{512}$ for the $j$th edge in the topological map.
    \item A single global feature $\mathbf{u}^{(k)} \in \mathbb{R}^{512}$.
    
\end{itemize}

\paragraph{Outputs of last GN block.} The last GN block (index $m$) uses the same input dimension as the intermediate GN blocks. For the output dimensions, we use the following:
\begin{itemize}
    \item Node features $\mathbf{v}^{(m)}_i \in \mathbb{R}^{768}$ for the $i$th node in the topological map.
\end{itemize}



\subsubsection{Cross-Modal Transformer}

\paragraph{Architecture.} Our cross-modal transformer uses a similar architecture to LXMERT~\cite{tan-bansal-2019-lxmert}. While LXMERT uses object-level image embeddings which are encoded with object position information via extra layers, we use node embeddings (from the GNN) and encode position using learnable position embeddings as described in Sec.~\ref{sec:method-planner-architecture}.

For the language encoding branch, we map each word to a learnable embedding and each index position to a learnable embedding as well. For words $w_i$ at index $i$ of a navigation instruction, we encode the instructions as:
\begin{align}
    h_i &= \text{LayerNorm}(\text{WordEmbed}(w_i) + \text{IdxEmbed}(i))
\end{align}

\paragraph{Single-modality encoders.} Each layer of the single-modal encoders contains a self-attention sub-layer followed by a feed-forward sub-layer. Each sub-layer is followed by a residual connection~\cite{he2016deep} and layer normalization~\cite{ba2016layer}.

\paragraph{Cross-modality encoders.} Each layer of the cross-modality encoder contains a cross-attention sub-layer, a self-attention sub-layer, and a feed-forward sub-layer. Each sub-layer is followed by a residual connection~\cite{he2016deep} and layer normalization~\cite{ba2016layer}. 

\paragraph{Number of layers.} In total, we use 9 single-modal language encoding layers, 5 single-modal map encoder layers, and 5 cross-modal layers. For further details, we refer the reader to Tan and Bansal~\cite{tan-bansal-2019-lxmert}.

\paragraph{Classification head.} We use a linear layer in the classification head to map from the 768-dimensional transformer node outputs to the classification logits.

\paragraph{Loss function.} We use a standard cross-entropy loss for training the cross-modal transformer planner (CMTP), where classification is done over the nodes and \verb|[STOP]| action. The same loss is used for CMTP-Repeated (CMTP-R) except the classification is done only over the nodes. For CMTP Binary Cross Entropy (CMTP-BCE), we use the following loss function:
\begin{align}
    L = 0.5 \text{CE}(x_{node}, y_{node}) + 0.5 \text{BCE}(x_{stop}, y_{stop})
\end{align}
where $CE$ represents cross-entropy loss over the node predictions $x_{node}$ and ground truth $y_{node}$, and $BCE$ is binary cross entropy over the predicted stop action $x_{stop}$ and ground truth $y_{stop}$.

\paragraph{Training details.} For the training the planner, we use the AdamW optimizer~\cite{kingma2014adam,loshchilov2018fixing} with a learning rate of 2e-5 and a linear warm-up schedule. We use a weight decay of 0.01 on all non-bias and layer normalization~\cite{ba2016layer} weights. We use a batch size of 16. All planners (GNN baseline and CMTP models) are trained from scratch without pre-training.

\subsection{Controller}

\subsubsection{Logic}

The controller is given a navigation plan in the form of a path in the topological map. It must then execute the plan by using the observations to predict an action at each time step. Specifically, Alg.~\ref{alg:control-alg} describes the logic in which the controller interacts with the environment. 

\begin{algorithm}
\SetAlgoLined
\SetKwInOut{Input}{input}
\SetKwInOut{Hyperparam}{param}
\Input{$\{o_1,o_2,...,o_n\}$ panoramas at planned nodes }
\Hyperparam{$\sigma^2$ variance of bias function \newline $d$ localization threshold}
 $\phi'_0 \leftarrow \vec 1, k \leftarrow 1, t \leftarrow 1$, \sc{observe} $o_0$\\
 \While{$k \le n$}{
    \sc{observe} $o_{curr}$\\
    $\phi_t,\rho_t \leftarrow C_{high}(o_{k-1}, o_{curr}, o_k)$ \\
    \eIf{$\rho_t \le d$}{
        $k \leftarrow k + 1$
    }{
        $\phi'_t \leftarrow B (\phi_t, \phi'_{t-1}; \sigma^2)$ \\
        \sc{execute} $C_{low}(\arg\max_i \phi'_{t_i})$ \\
        $t \leftarrow t+ 1$
    }
 }
 \caption{Controller Logic}
 \label{alg:control-alg}
\end{algorithm}

\begin{figure*}
    \centering
\includegraphics[width=.78\textwidth]{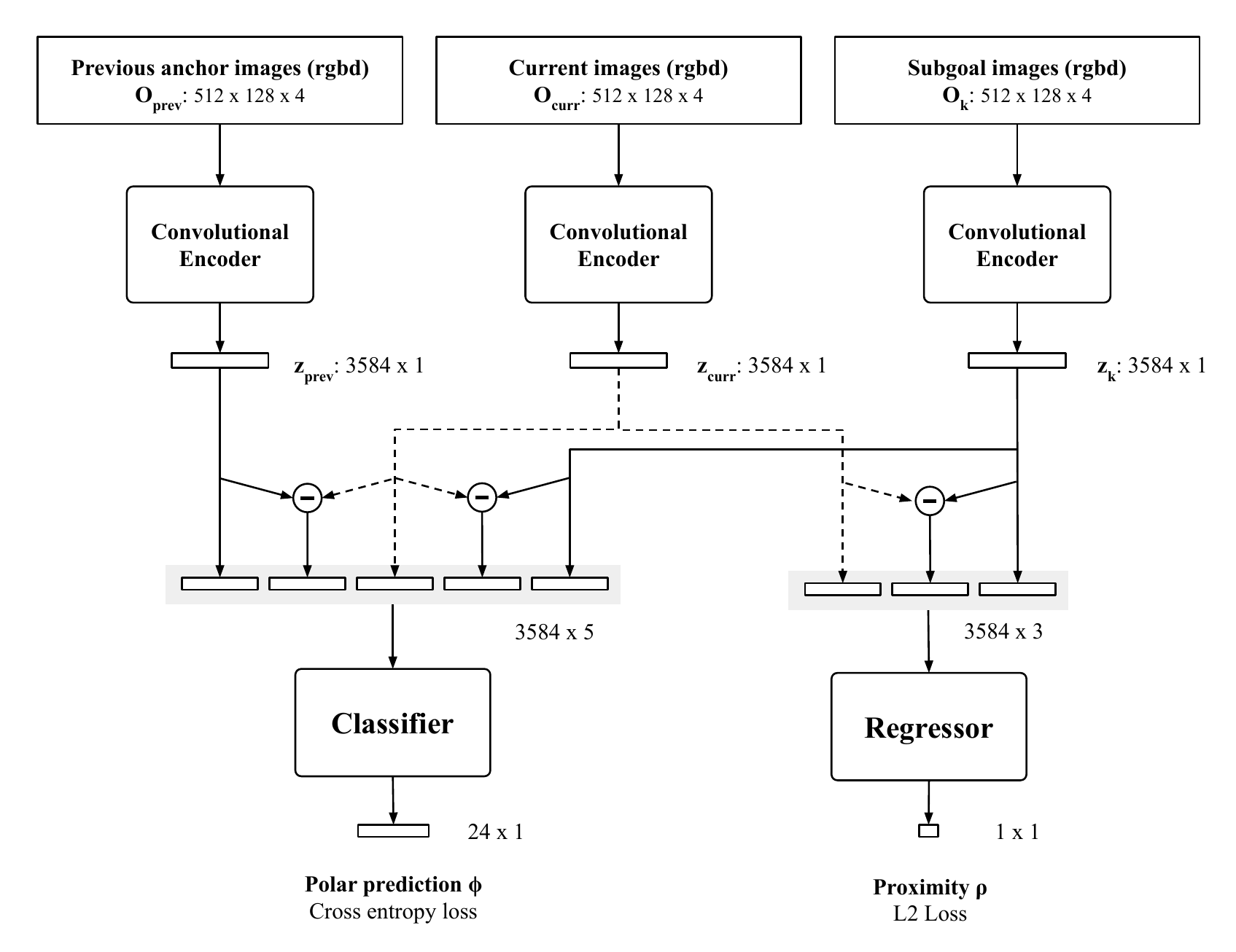}
    \caption{\textbf{High-level controller architecture.} The high-level controller is composed of three main components: the convolutional encoder, classifier, and and regressor. See the text for details.}
    \label{fig:controller-arch}
\end{figure*}

The controller first retrieves the RGBD panoramic observation for each node in the navigation plan. At the beginning of an episode, the controller assumes that the agent is in proximity to the first node $o_1$ in the plan. It then selects $o_2$ as the subgoal. At each time step, the agent makes an observation of the environment $o_{curr}$, and uses an odometry sensor reading of the agent's orientation to rotate $o_{curr}$ to align with the orientation of $o_1$ and $o_2$. 

$C_{high}$ predicts a waypoint based on these three panoramas, a bias function $B$ is applied to the prediction to alleviate perceptual aliasing, and then $C_{low}$ translates that waypoint into a robotic action. If the predicted waypoint is within a hyperparameterized proximity threshold such that $\rho \le d$, the agent consumes the current subgoal, and the next node in the plan becomes the new subgoal. This process is repeated until there are no more nodes in the plan. 

\subsubsection{High level controller ($C_{high}$)}

The high level controller, similar to \cite{meng2019scaling}, is a robot-agnostic function that predicts a waypoint when given an agent state. As specified above, the agent state is denote by an RGBD panoramic observation triplet $<o_{prev}, o_{curr}, o_{subgoal}>$. The output of the high level controller is a waypoint, towards which the agent should move to in order to reach the position of $o_{subgoal}$. This waypoint is denoted in a discretized polar coordinate system defined by $<\phi, \rho>$, where $\phi$ is the index of one of 24 partitions of each spanning 15 degree, and $\rho$ is a numeric distance.  

As such, the high level controller is a function that maps the agent state of observations to a waypoint: 
\begin{align}
    C_{high}(o_{prev}, o_{curr}, o_{subgoal}) \rightarrow \hat\phi ,\hat\rho
\end{align}

\begin{table}[]
    \centering
    \begin{tabular}{|c|c|c|c|c|}
        \hline
         Layer & In & Out & Kernel & Stride \\
        \hline
         PConv & 4 & 64 & 7 & 3 \\ 
         Batch Norm & 64 & 64 & & \\
         ReLU & 64 & 64 & & \\
         Max Pool & 64 & 64 & 3 & 2\\
         \hline
         PConv & 64 & 128 & 5 & 2 \\ 
         Batch Norm & 128 & 128 & & \\
         ReLU & 128 & 128 & & \\
         Max Pool & 128 & 128 & 2 & 2\\
         \hline
         PConv & 128 & 256 & 3 & 1 \\ 
         Batch Norm & 256 & 256 & & \\
         ReLU & 256 & 256 & & \\
         Max Pool & 256 & 256 & 2 & 2\\
         \hline
         PConv & 256 & 512 & 3 & 1 \\ 
         Batch Norm & 512 & 512 & & \\
         ReLU & 512 & 512 & & \\
         Max Pool & 512 & 512 & 2 & 2\\
        \hline
        Flatten+FC & & 3584 & &\\
        \hline
    \end{tabular}
    \caption{\textbf{Convolutional encoder architecture.} Each row represents a layer in the network along with the number of input channels (In) and number of output channels (Out).}
    \label{tab:convnet_arch}
\end{table}

\begin{table}[]
    \centering
    \begin{tabular}{|c|c|c|}
        \hline
         Layer & Shape In & Shape Out \\
        \hline
        FC & 3584 x 5 & 512 \\ 
        ReLU & 512 & 512 \\
        \hline
        FC & 512 & 256 \\ 
        ReLU & 256 & 256 \\
        \hline
        FC & 256 & 128 \\ 
        ReLU & 128 & 128 \\
        \hline
        FC & 128 & 24 \\ 
        \hline
    \end{tabular}
    \caption{\textbf{Classifier architecture.} The classifier predicts the direction of the waypoint. See the text for details.}
    \label{tab:classifier_arch}
\end{table}

\begin{table}[]
    \centering
    \begin{tabular}{|c|c|c|}
        \hline
         Layer Type & shape in & shape out \\
        \hline
        FC & 3584 x 3 & 256 \\ 
        ReLU & 256 & 256 \\
        \hline
        FC & 256 & 128 \\ 
        ReLU & 128 & 128 \\
        \hline
        FC & 128 & 1 \\ 
        \hline
    \end{tabular}
    \caption{\textbf{Regressor architecture.} The regressor predicts the waypoint distance from the agent.}
    \label{tab:regressor_arch}
\end{table}

\paragraph{Architecture.} As seen in Figure \ref{fig:controller-arch}, our controller architecture is a general 2-staged approach. We share the same convolutional encoder to extract embeddings from each of our observations $O_{prev}$, $O_{curr}$ and $O_k$ as explained in the paper. These embeddings are then passed to the classifier and regressor. Note that the input to the classifier is a concatenated 5-tuple of $(z_{prev}, z_{prev} - z_{curr}, z_{curr}, z_{curr} - z_{k}, z_k)$, and the input to the regressor is a concatenated 3-tuple of $(z_{curr}, z_{curr} - z_{k}, z_k)$, where each $z$ is the 3584 dimensional output from the convolutional encoder. We find that including these ``deltas'' that are the difference between the embeddings is helpful for training a better controller. The architectures for the convolutional encoder, classifier, and regressor are depicted in Table~\ref{tab:convnet_arch}, Table~\ref{tab:classifier_arch}, and Table~\ref{tab:regressor_arch}, respectively.

\paragraph{Training.} We perform the training of the high level controller network offline using a synthetically generated dataset of 2 million data samples. Each data sample consists of $<o_{prev}, o_{curr}, o_{subgoal}>$ and its label  $<\phi, \rho>$. Both output values are simultaneously optimized, with the direction index $\phi$ optimized by cross-entropy (CE), and distance $\rho$ by mean squared error (MSE):
\begin{align}
    L = 0.5 \text{CE}(\phi, \hat \phi) + 0.5 \text{MSE}(\rho, \hat\rho)
\end{align}
During training, we also perform data augmentation by randomly rotating all observation panoramas and $\rho$ by a same random amount.

\paragraph{Expert Waypoint.} The expert waypoint is produced by ground truth positions on a traversibility map (a grid). We first build a traversal graph $G$ of this grid, with edges connecting neighboring positions weighted by their L2 distance plus a penalty if close to a wall. Then, given any agent position $p_{curr}$ and a subgoal position $p_{subgoal}$, we can calculate a lowest cost path $p = \{p_1, p_2, ..., p_n\}$, where $p_1 = p_{curr}$ and $p_n = p_{subgoal}$. 

To get an expert waypoint, we define a maximum lookahead distance $m$, and we take the position $p_k$ such that travelling from $p_1$ to $p_k$ via the path is as far as possible, but no farther than $m$. 
Note that in this setting, when the agent position is sufficiently close to the subgoal, $p_k = p_{subgoal}$ is chosen such that the expert waypoint position is the subgoal position.

$C_{high}$ is a simple function that produces a waypoint based on observations, which can be trained completely offline given a sufficiently large and robust dataset of samples. We also use DAgger \cite{ross2011reduction} to further improve on the initially collected dataset.

\subsubsection{Low level controller ($C_{low}$)}

We use a simple low-level controller $C_{low}$ that rotates the agent to orient towards the predicted waypoint and walk in a straight line towards it. Because of the naive nature of this low-level controller, the waypoint predicted by $C_{high}$ must be highly accurate in order to achieve good performance. Our experiments in Sec.~\ref{sec:experiments} indicate that despite using a simple controller, the agent is still able to successfully follow navigation plans with over 80\% success rate. This performance could be further improved by using more sophisticated low-level controllers without requiring substantial changes to the rest of the overall approach due to our modular setup. 

\section{Experiments}

\begin{figure*}
    \centering
\includegraphics[width=\linewidth]{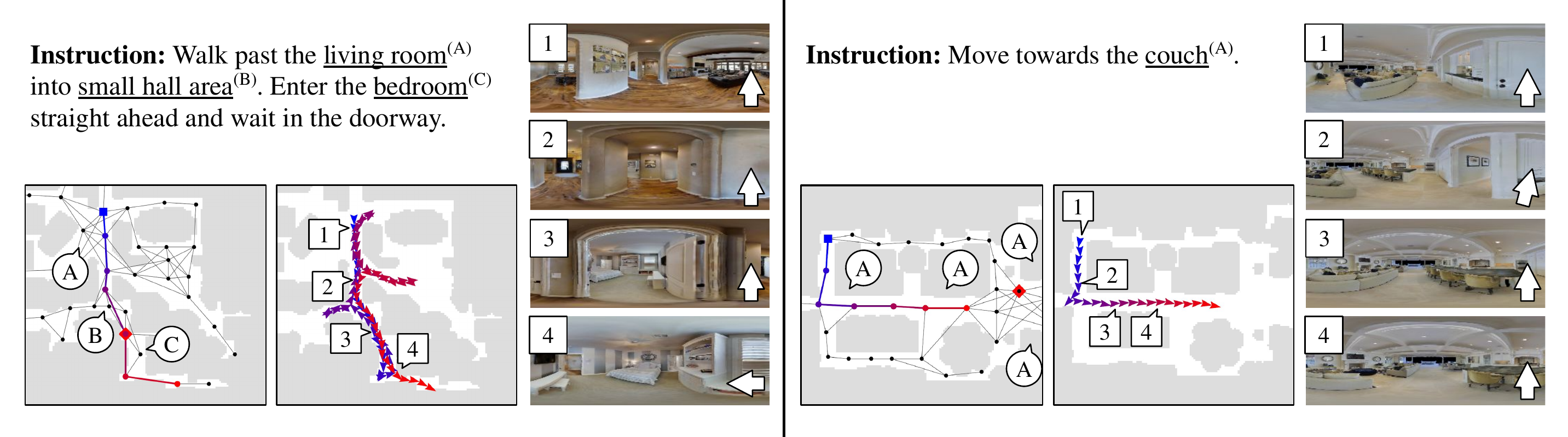}
    \caption{
        \textbf{Failure cases}. On the left, the planner overshoots the ground truth destination, failing to stop at the right time. The controller successfully navigates into the room and accidentally backtracks out. However, it still reaches the planner destination at the end of the episode. On the right, the navigation instruction is too ambiguous. There are many couches in the environment and it is unclear which couch to move towards. See the text for more details.}
    \label{fig:integration-failure}
\end{figure*}


We use the iGibson simulator~\cite{xia2019gibson,xia2020interactive} along with the VLN-CE data~\cite{krantz2020navgraph} for our experiments. To perform training and evaluation, we convert the VLN-CE trajectories into paths in our agent-generated (AG) maps and pre-defined Room2Room (R2R) graphs~\cite{anderson2018vision}. For each episode, we compared the ground truth 3D trajectory positions with the closest nodes in the appropriate topological maps. Episodes which contained start/goal positions that were $>2m$ from the closest topological map node were discarded for both train and validation splits. The baseline GNN and VLN-CE models were trained in iGibson~\cite{xia2020interactive}.


\subsection{VLN Qualitative Results}

\begin{figure*}
    \centering
\includegraphics[width=\linewidth]{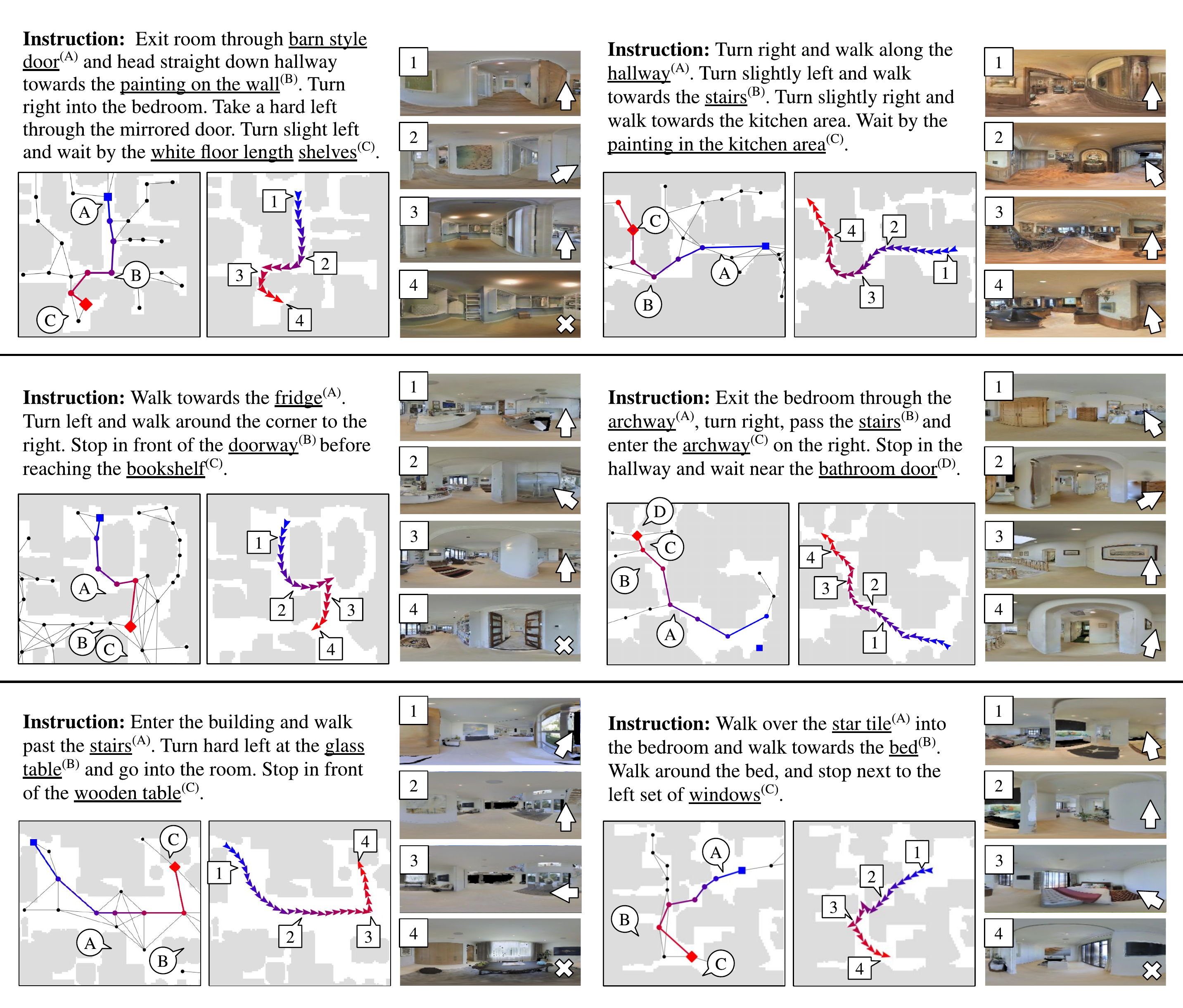}
    \caption{
        \textbf{Integrated system VLN examples.} We show 6 success cases of our CMTP approach. The agent successfully predicts plans according to the directions and landmarks specified in the instructions, while the controller executes the plans via low-level actions.}
    \label{fig:integration-success}
\end{figure*}

We provide more qualitative results of our integrated system (CMTP) in Fig.~\ref{fig:integration-success}. As can be seen in the figure, the planner successfully generates plans which land near the ground truth goal position. Many of the trajectories involve going into multiple rooms or navigating past landmarks specified in the instructions.

We also show example failures and inefficiencies in Fig.~\ref{fig:integration-failure}. On the left side of Fig.~\ref{fig:integration-failure}, we see the planner traverses past the ground truth goal position, failing to stop at the right time. On the other hand, the controller initially successfully navigates into the bedroom. However, it mistakenly backtracks out of the room and after numerous self-corrections, finally correctly navigates into the bedroom and stops at the end of the navigation plan.

In the example on the right of Fig.~\ref{fig:integration-failure}, the provided instruction is too ambiguous. The instruction tells the agent to move towards the couch, but there are several couches in the environment and it is unclear which couch is being referred to in the instruction. As a result, the planner predicts a plan that moves towards a couch and the controller successfully executes the navigation plan, but this is not the same couch as specified in the instruction.

\subsection{Cross Modal Attention}

\begin{figure*}
    \centering
\includegraphics[width=.9\linewidth]{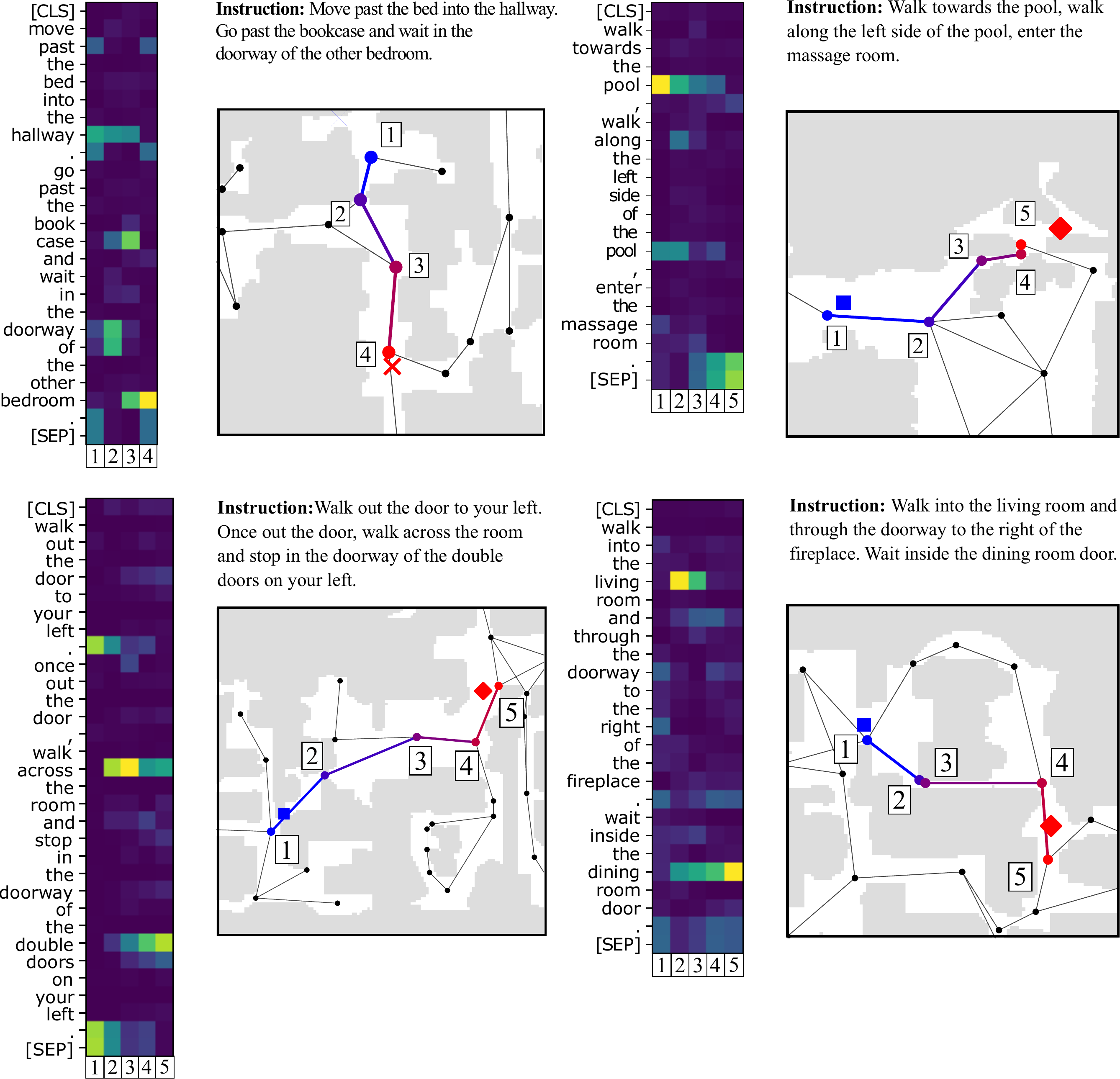}
    \caption{\textbf{Cross modal attention.} These visualizations follow the same format as Fig.~\ref{fig:xmodal-attention}. See the text for details.}
    \label{fig:xmodal_attn}
\end{figure*}

In Fig.~\ref{fig:xmodal_attn} we include additional examples of cross modal attention. These follow the same structure as Fig.~\ref{fig:xmodal-attention}. In the top-left example, we see ``hallway'' associated with nodes in the hallway, and the ``bedroom'' word associated with the bedroom node (node 4).

Similarly, in the bottom-left example, we see that the word ``across'' has high correlation with the nodes involved in traversing across the room (nodes 2, 3, 4), while ``double doors'' is related to the last node.  These results suggest that the model is capable of aligning instruction words with spatial locations.

\end{document}